\documentclass[a4paper, 12pt] 
{rsproca_new}

\usepackage{color} 
\usepackage{bm}  

\usepackage{tcolorbox}

\titlehead{Research}

\begin{document}

\title{Numerical solutions of fixed points in two-dimensional Kuramoto-Sivashinsky equation expedited by reinforcement learning}

\author{
Juncheng Jiang$^{1}$, Dongdong Wan$^{1}$ and Mengqi Zhang$^{1}$}

\address{$^{1}$Department of Mechanical Engineering, National University of Singapore, Singapore}

\subject{artificial intelligence, fluid mechanics}

\keywords{deep reinforcement learning, Kuramoto–Sivashinsky equation, fixed point}

\corres{Mengqi Zhang\\
\email{mpezmq@nus.edu.sg}}

\begin{abstract}
This paper presents a combined approach to enhancing the effectiveness of Jacobian-Free Newton-Krylov (JFNK) method by deep reinforcement learning (DRL) in identifying fixed points within the 2D Kuramoto-Sivashinsky Equation (KSE). JFNK approach entails a good initial guess for improved convergence when searching for fixed points. With a properly defined reward function, we utilise DRL as a preliminary step to enhance the initial guess in the converging process. The main advantage brought about by the reward function in DRL is to identify potential initial guess candidates with similar spectral structures over time, which facilitates the search of fixed points. We report new results of fixed points in the 2D KSE which have not been reported in the literature. Additionally, we explored control optimization for the 2D KSE to navigate the system trajectories between known fixed points, based on parallel reinforcement learning techniques. This combined method underscores the improved JFNK approach to finding new fixed-point solutions within the context of 2D KSE, which may be instructive for other high-dimensional dynamical systems.
\end{abstract}


\maketitle

\section{Introduction}

The Kuramoto–Sivashinsky equation (KSE) represents one of the simplest nonlinear systems, which exhibits complex spatio-temporal dynamics \cite{KSE_bridge,Cvitanovic2010}. It was derived by Kuramoto \cite{Kuramoto} in angular-phase turbulence for a system of reaction-diffusion equations, and by Sivashinsky \cite{Sivashinsk} to model small thermal diffusive instabilities in laminar flame fronts. Beyond its inherent physical relevance, KSE has also garnered substantial mathematical interest. It has become an important model for investigating the complex dynamics of chaotic systems. 

To understand the dynamics and orbital behaviours of chaotic systems, the role of fixed points stands paramount \cite{Strogatz2014}. Herein, an orbit is defined as the temporal evolution of an initial condition within a function space. When a fixed point exhibits stability, nearby orbits inherently gravitate towards it, categorising it as an attractor. In contrast, the presence of an unstable fixed point leads to the divergence of proximal orbits. When all fixed points are unstable, the orbit may engage in a cyclical movement among the fixed points. Yet, if these recurrent orbits similarly display instability, the orbits could potentially navigate indefinitely within the function space, ultimately leading to convergence on a strange attractor. This viewpoint has been a plausible theory to understand more advanced topics such as turbulence \cite{Lanford1982,Kerswell2005,Kawahara2011,Graham2021}. Hence, pinpointing fixed points remains crucial in unraveling the complexities of chaotic systems, such as those in the KSE framework. 

The fixed points of 1D KSE have received substantial interest among researchers, primarily due to their pivotal role in elucidating the onset of chaos and pattern formation in nonlinear dynamical systems \cite{tadmor1986, smyrlis1991, kevrekidis1990, otto2019} among many others. The inquiry into the steady solutions of 1D KSE, as examined by Greene \& Kim \cite{Steady_State}, constitutes a significant advancement in the field. In a subsequent study, Lan \& Cvitanović \cite{lan2008} applied the Newton descent method to accurately identify the unstable fixed points of the 1D KSE, thereby deepening the understanding of the complexities of the system. In addition, Cvitanović \textit{et al.} \cite{Cvitanovic2010} mapped out the spatial distributions of these fixed points for the domain length $L = 22$, offering a detailed perspective on their spatial dynamics. Collectively, these works illuminate the stability and bifurcation properties inherent in the 1D KSE, significantly advancing our understanding of the intricate dynamics characteristic of chaotic systems.

The Jacobian-Free Newton-Krylov (JFNK) method stands as a classical numerical approach employed for calculating and identifying fixed points in partial differential equations (PDEs). This methodology obviates the requirement for explicit formation of the Jacobian matrix, showcasing extensive applicability \cite{JFNK}, particularly suitable for large-scale numerical problems. However, when addressing intricate nonlinear PDEs, the convergence of JFNK is contingent upon good initial guesses. In instances of suboptimal initial guesses, JFNK might experience convergence failures. Consequently, when confronted with complex chaotic systems such as 2D KSE, a good initial guesses or effective preconditioning strategies becomes important.

In recent years, deep reinforcement learning (DRL), a subfield of artificial intelligence, has attracted substantial attention due to its capacity to acquire optimal control policies by interacting with a dynamic environment. Leveraging its data-driven characteristics and adaptability, DRL has demonstrated remarkable efficiency in accomplishing tasks across various domains, including robotics \cite{Robotics}, natural language processing \cite{Language},  multiple games \cite{Mutiple_Game}, AlphaGo \cite{AlphaGo}, and autonomous vehicles \cite{Automatical_Drive}.
DRL's outstanding performance in various fields and its ability to solve complex, nonlinear problems have also sparked interest in applying DRL to fluid mechanics applications \cite{Brunton2020,DRL_review2}. In recent years, efforts to apply DRL to fluid mechanics have been ongoing. In areas such as drag reduction \cite{Drag,Paris2021,Li2022,Xu2023,Sonoda2023}, heat transfer \cite{Heat}, shape optimization \cite{Shape_Optimization}, and flow control \cite{Flow_Control2,zeng2021,chaotic} among many others, DRL has achieved remarkable accomplishments.

In the research presented by Bucci \textit{et al.} \cite{chaotic}, DRL was successfully utilized to stabilize the dynamics of the 1D KSE around its inherently unstable fixed points. Concurrently, another study by Zeng \& Graham \cite{zeng2021} demonstrated the capability of DRL in identifying the fixed points of the 1D KSE. Building upon these foundational works, this research aims to extend these methodologies and insights to the 2D KSE framework.
Transitioning from the 1D to the 2D KSE imposes substantial complexities when employing DRL for system control, predominantly due to the augmented degrees of freedom and the intricate spatial-temporal interdependencies inherent in a 2D domain. {\color{black}The 2D formulation requires significantly more computational resources because it needs denser spatial discretisation in numerical simulations. This increases the complexity of DRL’s computational demands. } Coupled with the elaborate interaction dynamics between different spatial modes in 2D, there are more unpredictable and multifarious system behaviours that the DRL must learn to decipher and adapt to. 

To summarise the literature review, the augmented dimensionality in a complex dynamical system presents challenges for identifying its fixed points. The conventional solution method based on JFNK may encounter convergence issues when identifying fixed points in high-dimensional dynamical systems, especially when lacking robust preconditioning techniques or properly selected initial guesses. Based on a combined method of JFNK and DRL, we report in this work computational efforts in stabilising the dynamics of the 2D KSE converging to its unstable fixed points. The integrative method leverages DRL to generate good initial guesses. Subsequent to this initialisation, the JFNK method is employed for identifying the fixed points, facilitating a streamlined solution process. As a result, more than 300 fixed points in 2D KSE have been found and listed in section \ref{fixedpoints}.

The paper is organized as follows. Section 2 explains the foundational concepts pertinent to the current study. In section 3, we provide an in-depth exposition of our methodologies, which include the control of the 2D KSE, DDPG algorithm, {\color{black} the generation method for initial guesses}, DRL reward and the exploration noise in DRL. Section 4 presents the result of numerical simulations, fixed points of 2D KSE and DRL-based navigation between fixed points. Finally, we summarize our findings in section 5. The appendices explains the convergence issue in the JFNK method and the fine-tuning of the DRL framework, including the Bayesian optimisation of hyperparameters and the exploration noise optimisation. {\color{black}We have made our code public at https://github.com/Jiang-JC/2D-KSE-RL.}

\section{Problem formulation}
\subsection{Two-dimensional Kuramoto-Sivashinsky equation}

The KSE has a general form that can be expressed as a nonlinear, fourth-order partial differential equation, given by
\begin{equation}
\phi_t + \frac{1}{2} |\nabla \phi|^2 + \Delta {\phi}  + \Delta ^ 2 {\phi} = F({\mathbf x},t).
\end{equation}
where $\phi$ is the dependent variable that describes the behavior of the system over time and space. The term $F({\mathbf x},t)$, to be explained below, is the control force which will be determined by DRL in our combined method. In a 2D case, the KSE can be explicitly written as
\begin{equation}
{\phi}_t + \frac{1}{2} ({{\phi}_x^2} + {{\phi}_y^2}) + {\phi}_{xx} + {\phi}_{yy} + {\phi}_{xxxx} + 2{\phi}_{xxyy}  + {\phi}_{yyyy} = F(x,y,t), \label{2DKSE}
\end{equation}
where $x$, $y$, and $t$ are the spatial and temporal coordinates, respectively. The variables $\phi_x$,  $\phi_y$ and $\phi_t$ denote the partial derivatives of $\phi$ with respect to $x,y,t$. The computational domain is $[0,2L]\times[0,2L]$ with $2L=20$ in the $x,y$ directions and periodic boundary conditions are assumed.


In the context of PDEs, a fixed point refers to a solution or a point that remains unchanged after applying a certain transformation or operation. More formally, we denote the trajectory of $\bm{u}_t$ as the state of the 2D KSE at a given time $t$. This trajectory can be followed using the flow-map represented as $\bm{\Phi}^t$. The map accepts an initial point $u_0$ and advances it through a temporal interval $\Delta t$, such that $\bm{\Phi}^{\Delta t}: \bm{u}_0 \rightarrow \bm{u}_{\Delta t}$. An equilibrium $\bm{u}^*$ constitutes a fixed point that satisfies $\bm{u}^* = \bm{\Phi}^{\Delta t}(\bm{u}^*)$ for any temporal interval $\Delta t$.


%

\section{Numerical methods}

\subsection{Numerical simulations and control setup for 2D KSE}

To control the behavior of the 2D KSE and stabilise its unstable fixed points, the actuators and sensors are introduced into the system. These actuators and sensors are uniformly spaced along the $x$-axis and $y$-axis, motivated by the works of \cite{chaotic,zeng2021} in 1D KSE. The actuator is Gaussian shaped, and the forcing formula for $F(x,y,t)$ in equation (\ref{2DKSE}) is given by
\begin{equation}
F(x,y,t) = {\sum_{i=1}^{m}}{\sum_{j=1}^{m}}u_{ij}(t)\frac{1}{2\pi{\sigma}^2}{\exp}{(-\frac{(x-{x_{i}^a})^2+(y-{y_{j}^a})^2}{2\sigma^2})}
\end{equation}
where the external force distributions parameter $m$ represents the number of actuators in each row (or column), while the standard deviation $\sigma$ defines the spatial distribution of each actuator. The amplitude of the component in the $i$th row and $j$th column of the action vector is denoted by $u_{ij}(t)$. 

\begin{figure}
  \centering
   \includegraphics[width=1.0\textwidth]{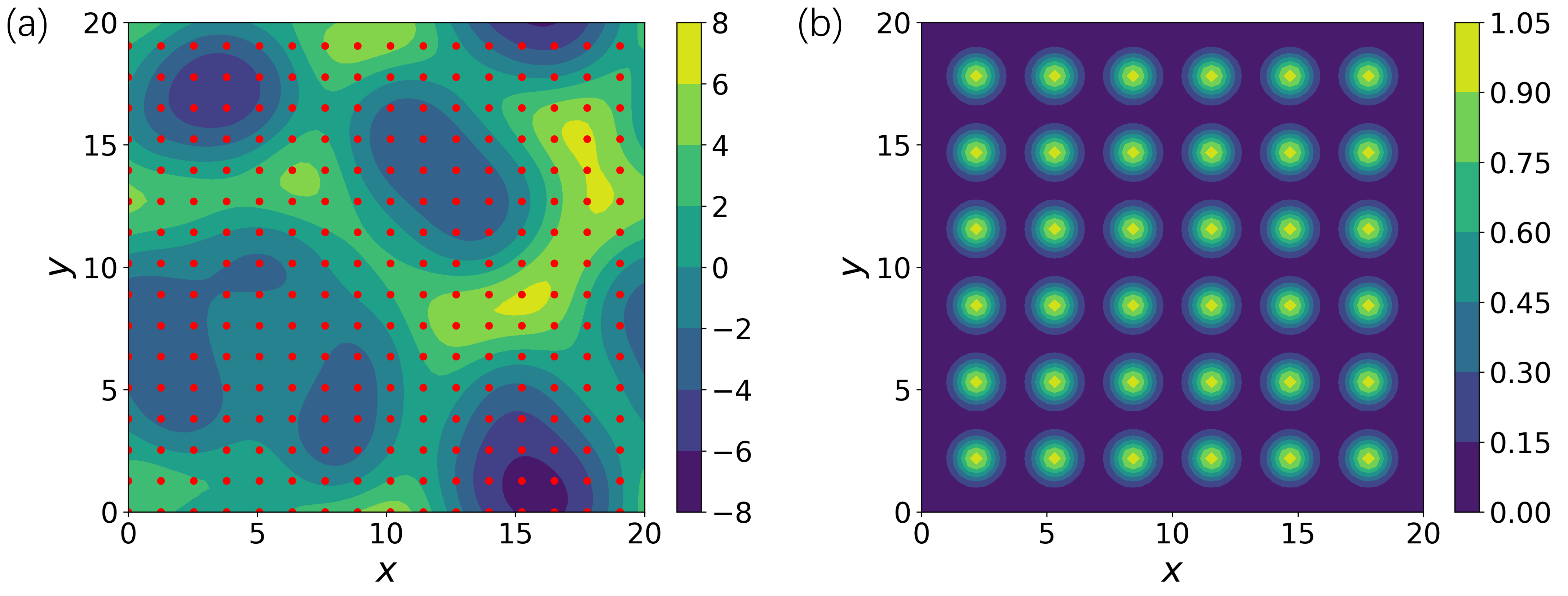}
  \caption{Sensor distribution (a) and force distribution (b) in the domain of 2D KSE. The color in panel (a) is dependent variable $\phi$ and red points denote sensors. In panel (b), the color represents the external force term of the 2D KSE. For enhanced visualization, we have set $\sigma=0.6$ and standardized all action vector amplitudes $u_{ij}=1$.}
      \label{Fig:Force_External}

\end{figure}

Figure \ref{Fig:Force_External} illustrates the configuration of sensors (red dots) in panel (a) and actuators in panel (b). With regard to the sensors, it is assumed that practical controllers utilize only partial information, reflecting real-life constraints. There are $16\times16=256$ sensors in place that measure the local velocity, with their positions being uniformly spaced with a spacing of $\Delta x = \Delta y = 2L/16$. Specifically, the sensor locations are defined as $x_i^s{\in}\{0, 4, 8, 16...60\}2L_x/64$, $y_i^s{\in}\{0, 4, 8, 16...60\}2L_y/64$. For the actuators, through the parameter optimization detailed in Appendix B, it is determined that the configuration with $m=6$ and $\sigma=2.4$ yields the most efficacious results in DRL applications, leading to a total of 36 actuators, as shown in panel (b). The locations of the actuators along the $x$-axis and $y$-axis are denoted as  ${x_i^a}{\in}\{8,18,28,38,48,58\}2L_x/64$ and ${y_i^a}{\in}\{8,18,28,38,48,58\}2L_y/64$, respectively.

To perform numerical simulations of 2D KSE, we utilize a spectral method for the spatial discretization and an exponential time-differencing algorithm along with a RK4 time-advancement scheme, developed from the corresponding 1D version presented in \cite{Numerical}. We impose periodic boundary conditions expressed as $\phi(x, y, t) = \phi(x + 2L, y, t) = \phi(x, y + 2L, t)$. The inherent spatial periodicity facilitates the projection of the instantaneous solution onto distinct Fourier modes. For a domain length of $2L=20$, we employ $N = 64 * 64$ Fourier collocation points for the spatial discretisation. For all numerical computations, the time increment is equal to 0.05.

To validate our configuration for Fourier collocation points and time increments, we conduct both the grid independence check and a time-step independence check. Under the condition of maintaining all other variables constant, we conduct the numerical tests spanning 25 time units, equivalent to 500 time steps. In the grid independence check, we compare results obtained using $N = 64 * 64$ and $N = 128 * 128$ Fourier collocation points. The outcomes were nearly identical, exhibiting an error margin smaller than $10^{-6}$. Regarding the time-step independence check, comparisons were made between simulations with time increments of 0.05 and 0.025. These simulations yielded closely aligned results, with most of the results showing an error of $10^{-4}$ and the maximum error being less than ${10}^{-2}$. This convergence study verifies our implementation of the numerical methods.  

\begin{figure}
  \centering
   \includegraphics[width=1.0\textwidth]{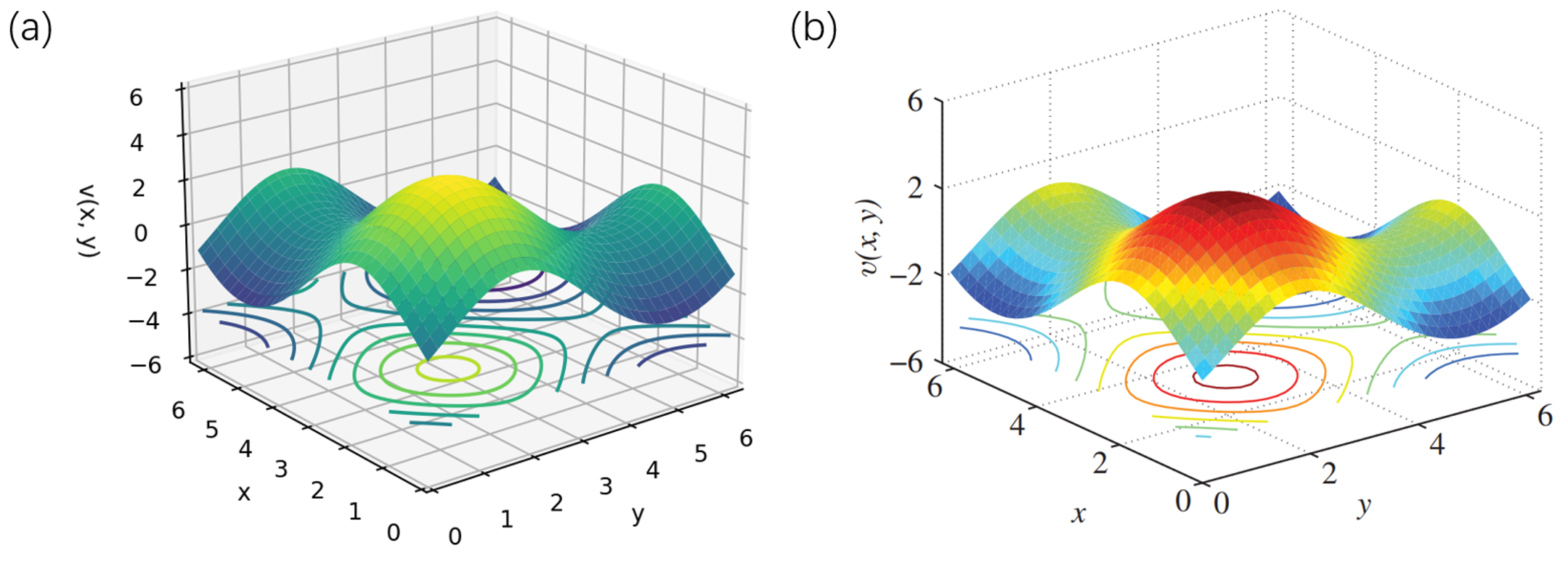}
  \caption{The steady state of the 2D KSE. (a) the result generated by our numerical code. (b) the result reported in Kalogirou \textit{et al.} \cite{2DKSE}. In both figures, the $x$-axis and $y$-axis span a range of $2\pi$, while the $z$-axis represents the dependent variable, which is denoted as $v$ in the reference.}
      \label{Fig:Valid1}
\end{figure}

In order to validate the results of our 2D KSE numerical approach against those in the existing literature, we conduct a comparative analysis between our numerical results and those found in the figure from \cite{2DKSE}, see figure \ref{Fig:Valid1}. The visualizations generated by our code align closely with the figure presented in the reference, thus serving as an effective validation of the accuracy and reliability of our simulation code. 

{\color{black}Finally, our JFNK implementation was adapted from the code provided in Willis \cite{JFNK_code}. The parameters used are maximum GMRES iterations $m_{gmres} = 100$, maximum Newton iterations $n_{its} = 100$, relative error tolerance $\varepsilon_{err} = 10^{-12}$, minimum trust region size $\delta_{min} = 10^{-20}$, maximum trust region size $\delta_{max} = 10^{20}$, tolerance for gradient $g_{tol} = 10^{-3}$, Jacobian approximation parameter $\varepsilon_j= 10^{-6}$, number of time steps $n_{dts} = 20$}. 

\subsection{Methodology and implementation of the DRL method}\label{method}
In the domain of DRL algorithms, we have adopted the Deep Deterministic Policy Gradient (DDPG) algorithm as our principal learning methodology. DDPG is a model-free, off-policy algorithm for continuous control problems, which combines ideas from both reinforcement learning and deep learning. The algorithm is a variant of the standard DPG (Deterministic Policy Gradient) algorithm and extends it to handle high-dimensional state and action spaces \cite{DDPG1}. 

DDPG uses an actor-critic architecture, where the actor is a deep neural network that learns a deterministic policy mapping from states to actions, and the critic is another deep neural network (NN) that learns the state-action value function. The actor is updated using the deterministic policy gradient, which is the gradient of the state-action value function with respect to the actor parameters. The critic is updated using the TD (Temporal Difference) learning algorithm, which estimates the state-action value function from the observed transitions in the environment \cite{DDPG2}. 

\begin{tcolorbox}[colback=green!5!white, colframe=green!50!black, title = Algorithm 1: DDPG for Identification of the Fixed Points] 
\textbf{KS environment related sensor and actuator hyparameters:} \\
$N_{SAMPLE}$ — the number of sample points , $S_{DIM}$ — the number of equispaced sensors, $A_{MAX}$ — the maximum amplitude for the actuation inputs, $m$  — the external force distribution parameter,  $\sigma$ — standard deviation of the Gaussian function\\

\textbf{DRL related hyperparameters:} \\
$B$ — batch size, $\alpha$ — learning rate, $\gamma$ — discount factor, $\tau$ — soft update rate for the target networks, $K$ — target network update frequency, $n_p$ — the number of parallel agents,  $\theta_{th}$ — reward threshold\\

\textbf{DRL related parameters:}\\ 
$Q$ — critic network, $\mu$ — actor network, $\mathcal{N}$ — exploration noise\\

Initialize critic network $Q(s, a | \theta^Q)$ and actor network $\mu(s | \theta^\mu)$ with weights $\theta^Q$ and $\theta^\mu$ \\
Initialize target networks $Q'$ and $\mu'$ with weights $\theta^{Q'} \leftarrow \theta^Q$, $\theta^{\mu'} \leftarrow \theta^\mu$ \\
Initialize replay buffer $R$ \\
\textbf{for} episode = 1, $M$ \textbf{do}

\quad Initialize a random process $\mathcal{N}$ for action exploration 

\quad Receive initial observation state $s_1$ 

\quad Initialize the state with maximum reward $s_{r_{max}} = s_1$

\quad \textbf{for} t = 1, $T$ \textbf{do}

\quad \quad Select action $a_t = \mu(s_t | \theta^\mu) + \mathcal{N}_t$ according to the current policy and exploration noise

\quad \quad Execute action $a_t$ and observe reward $r_t$ and next state $s_{t+1}$

\quad \quad Update the maximum reward and the state with maximum reward:            $s_{r_{max}} = s_t, r_{max} = r_t (\text{if} \ r_{max} < r_t$)

\quad \quad Store transition $(s_t, a_t, r_t, s_{t+1})$ in replay buffer $R$

\quad \quad Sample a random minibatch of $N$ transitions $(s_i, a_i, r_i, s_{i+1})$ from $R$

\quad \quad Set $y_i = r_i + \gamma Q'(s_{i+1}, \mu'(s_{i+1} | \theta^{\mu'}) | \theta^{Q'})$

\quad \quad Update critic by minimizing the loss: 
    \[ L = \frac{1}{N} \sum_{i} \left( y_i - Q(s_i, a_i | \theta^Q) \right)^2 \]
\quad \quad Update the actor policy using the sampled policy gradient: 
    
    \[\nabla_{\theta^\mu} J \approx \frac{1}{N} \sum_{i} \nabla_a Q(s, a | \theta^Q) |_{s=s_i, a=\mu(s_i)} \nabla_{\theta^\mu} \mu(s | \theta^\mu) |_{s=s_i} \]
    
\quad \quad Update the target networks: 
    \[ \theta^{Q'} \leftarrow \tau \theta^Q + (1 - \tau) \theta^{Q'}\]\[\theta^{\mu'} \leftarrow \tau \theta^\mu + (1 - \tau) \theta^{\mu'} \]
\quad Pass the state with the maximum reward $s_{r_{max}}$ to JFNK, if the maximum reward $r_{max} > \theta_{th}$
\end{tcolorbox}

{\color{black}
The primary parameters and training process of the DRL model are presented in Algorithm 1. It should be noted that the pseudocode is specifically designed for the task of identifying fixed points. The task of navigating towards a goal fixed point is comparatively simpler and involves slight differences, so this algorithm also remains a valuable reference for that task as well. The parameters of the DRL model are categorized into different sections and are summarised below.
\begin{itemize}
\item The first section consists of KS environment-related sensor and actuator parameters, including the number of sample points ($N_{SAMPLE} = 64$), the number of equispaced sensors ($S_{DIM} = 16 \times 16$), the external force distribution parameter ($m$), the standard deviation of the Gaussian function ($\sigma = 2.4$), and the maximum amplitude for the actuation inputs ($A_{MAX} = 3$). 

A higher $S_{DIM}$ increases precision but also adds computational cost, with 16 representing an effective balance between the two. As introduced earlier, the parameters $m$ and $\sigma$ are key components that define the structure of the actuation inputs and are interdependent. The maximum amplitude of actuation inputs, $A_{MAX}$, can be easily adjusted by experimenting with different values. 

\item The second section addresses DRL-related hyperparameters, including batch size ($B = 200$), learning rate ($\alpha = 0.001$), discount factor ($\gamma = 0.99$), soft update rate for the target networks ($\tau = 0.001$), target network update frequency ($K = 1$), and the Adam optimizer. These hyperparameters ($B$, $\alpha$, $\gamma$, $\tau$, $K$) are standard for the DDPG algorithm. Following numerical tests, we decided to maintain the values from Ref. \cite{chaotic} in 1D KSE due to their proven effectiveness. Additionally, the number of parallel agents ($n_p$) helps increase the diversity of transitions in the replay buffer and accelerates convergence, though it also raises computational costs. After testing, $n_p = 10$ was found to offer a favorable balance. The reward threshold ($\theta_{th} = -45$) is a predefined criterion that dictates the transfer of the state to the JFNK method, contingent upon whether the obtained reward exceeds this threshold.

\item The final section covers DRL-related parameters, which are adjusted throughout the training process. These include the critic network ($Q$), the actor network ($\mu$), and action exploration noise ($\mathcal{N}$). The structures of both the critic network ($Q$) and the actor network ($\mu$) are introduced in Section 3(b) below, while the action exploration noise ($\mathcal{N}$) is detailed in Section 3(b)(ii).
\end{itemize}
}

\begin{figure}
  \centering
  \includegraphics[width=1\textwidth]{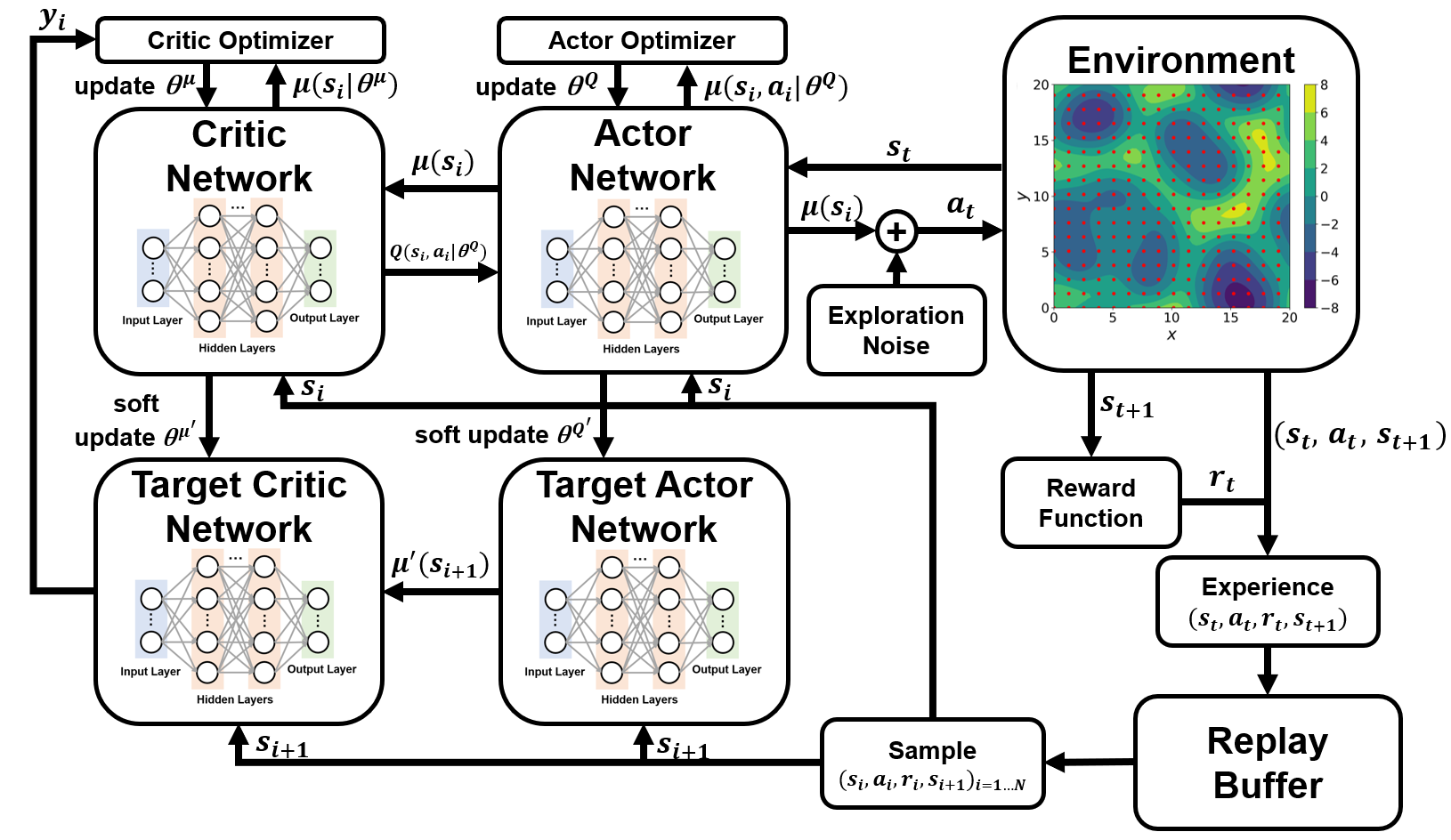}
  \caption{DRL schematic diagram. In the current DRL method, the environment is simulated by the 2D KSE. The state and action input are used to generate the subsequent state in the next time-step. The actor and target actor networks generate an action based on the current state. The critic and target critic networks evaluate the action's quality by estimating the Q-value based on the current state and action. This evaluation is then used to update the actor network. The replay buffer stores transitions of experiences that allow for efficient, batched updates of the actor and critic networks.}
       \label{Fig:DDPG}
\end{figure}

 
In the implementation aspect of the DDPG algorithm, we have utilized scripts crafted in PyTorch, adhering to the DDPG framework described in \cite{chaotic}. Figure \ref{Fig:DDPG} shows the configuration of the Critic and Actor networks. Pertaining to the critic's architecture,  the NN consists of an input layer of dimension 292, with 36 nodes earmarked for actuator signals, and 256 nodes allocated for sensor inputs. The output layer is a scalar, reflecting the resultant computed reward. Furthermore, this architecture incorporates two hidden layers containing 256 and 128 nodes respectively, both employing \texttt{swish} as the activation function. One of the key features of DDPG is the use of a replay buffer, which stores the transitions experienced by the agent in the environment. The replay buffer is used to decorrelate the data and improve the efficiency of the learning process. For the actor's architecture, the input layer, endowed with 256 nodes, is fed with the sensor input, whereas the output layer generates the requisite control signal feeding 36 actuators. This architecture also consists of two hidden layers, with node counts of 128 and 64. The activation functions applied are \texttt{swish} and \texttt{tanh}, respectively. Notably, the last layer functions with a saturation capability, wherein the amplitude of the output is constrained within the range of -3 to 3.

\subsubsection{Reward design in DRL} \label{reward_section}

In the context of reinforcement learning, reward is a fundamental concept that serves as a mechanism to guide an agent's learning process. It represents the numerical feedback provided to the agent by the environment after it takes a certain action in a specific state. The reward signal essentially quantifies the immediate benefit or desirability of the agent's action, aiding the agent in learning how to make decisions to maximize its long-term cumulative return.

Our study will consider two distinct tasks — navigation towards a goal fixed point and identification of fixed points. Correspondingly, we have established two separate rewards. For the first scenario, the reward is defined as the Euclidean distance between the current state of the system and the goal fixed point. This approach aligns with the reward employed in the work of 1D KSE conducted by \cite{chaotic}
\begin{equation}
    r := - \|u_t - u_g\|_2
\end{equation}
where $r$ represents the reward, $u_t$ the state of the current system and $u_g$ the state of goal fixed point which is known in this case.

For the second task to identify fixed points, we define the reward as the Euclidean distance in the spectral space between the state of the current system and its state after one time interval $\Delta t$ without the application of external forces
\begin{align}
r :=& - \|\text{FFT}({u_{t+\Delta t}}) - \text{FFT}(u_{t})\|_2, 
\end{align}
where \text{FFT} means fast Fourier transformation.

\subsubsection{Exploration noise in DRL}\label{noise}

In the study conducted by \cite{chaotic}, DRL was effectively employed to steer the 1D KSE towards the desired fixed points. However, when directly applying their approach to the 2D KSE, we found that the learning performance exhibited a substantial decline. In order to tackle this challenge, in addition to the parallel reinforcement learning approach (to be discussed further in Appendix B\ref{parallel_section}), we have also adjusted the exploration noise in the DRL framework.

In DRL, Gaussian noise is often incorporated as an exploration mechanism to promote diverse action selection. This noise, characterized by a Gaussian or normal distribution, introduces random variations to the policy outputs, ensuring that the agent does not prematurely converge to suboptimal policies. By perturbing actions with a stochastic component, the agent explores a broader state-action space, facilitating escape from local minima and enhancing generalization capabilities. Over time, as the agent gains more knowledge about the environment, the variance of the Gaussian noise can be decayed to allow a gradual transition from exploration to exploitation. In our study, Gaussian noise is sought in the form:
\begin{equation}
a = clip(\mu_\theta(s) + \epsilon, a_{Low}, a_{High}),
\end{equation}
\begin{equation}
\epsilon \sim \mathcal{N}(0, \alpha \times a_{Lim}), \ \ a_{Low} = -1.2a_{Lim}, \ \ a_{High} = 1.2a_{Lim},
\end{equation}
where $\theta$ is the parameter of the actor-network, $s$ is the observed state, $\mu_\theta(s)$ is the determined actor generated by the actor-network, $\epsilon$ is the exploration noise, which follows a Gaussian distribution. The clip function is used to limit actor between $a_{Low}$ and $a_{High}$. In addition, $a_{Lim}$ is a user-defined hyper-parameter and $\alpha$ is the exploration noise parameter.

In our investigation, within the DDPG framework, the exploration noise parameter is progressively reduced as the proficiency of the reinforcement learning agent consistently improves. This decremental process eventually stabilizes at a predetermined minimum noise threshold, denoted as $\alpha = \alpha_{min}$. This mechanism ensures that the noise magnitude does not become excessively small, which could otherwise limit DDPG's ability to explore uncharted behavioral trajectories. We observe that the magnitude of $\alpha_{min}$ profoundly impacts the efficacy of DRL. As a result, we conduct a systematic analysis of the parameter $\alpha_{min}$ in Appendix B\ref{noise_section}.

\section{Results and discussion} 

\subsection{Application of DRL-assisted JFNK iteration for the 2D KSE}

First, we explain how we integrate the DRL to provide a better initial guess for the JFNK method when solving for the fixed points in 2D KSE. Within the DRL-JFNK hybrid method, reinforcement learning possesses the capability to identify the initial guesses characterized by smaller relative residuals. Even though the DRL-assisted initial guesses are still not exactly the true fixed points, we can leverage these improved initial guesses with smaller residuals to kickstart the JFNK method. This advantage becomes more critical in high-dimensional systems.


The role of DRL agent in identifying initial guesses with smaller relative residuals is explained as follows. As explicated in Section \ref{method}\ref{reward_section}, the reward is formulated as the negative of the Euclidean distance between the state of the current system and its state after a time interval $\Delta t$ without the application of external forces. The DRL agent is programmed to persistently increase this reward. {\color{black}Thus, the DRL agent actively explores the state space of the 2D KSE to identify states that are close to fixed points. This process aims to maximize the rewards.} As a result, the agent possesses the capability to identify numerous points with smaller residuals within a single training process (as defined in the reward). Those points can serve as advantageous initial guesses for the JFNK method.



\begin{figure}
  \centering
  \includegraphics[width=1.0\textwidth]{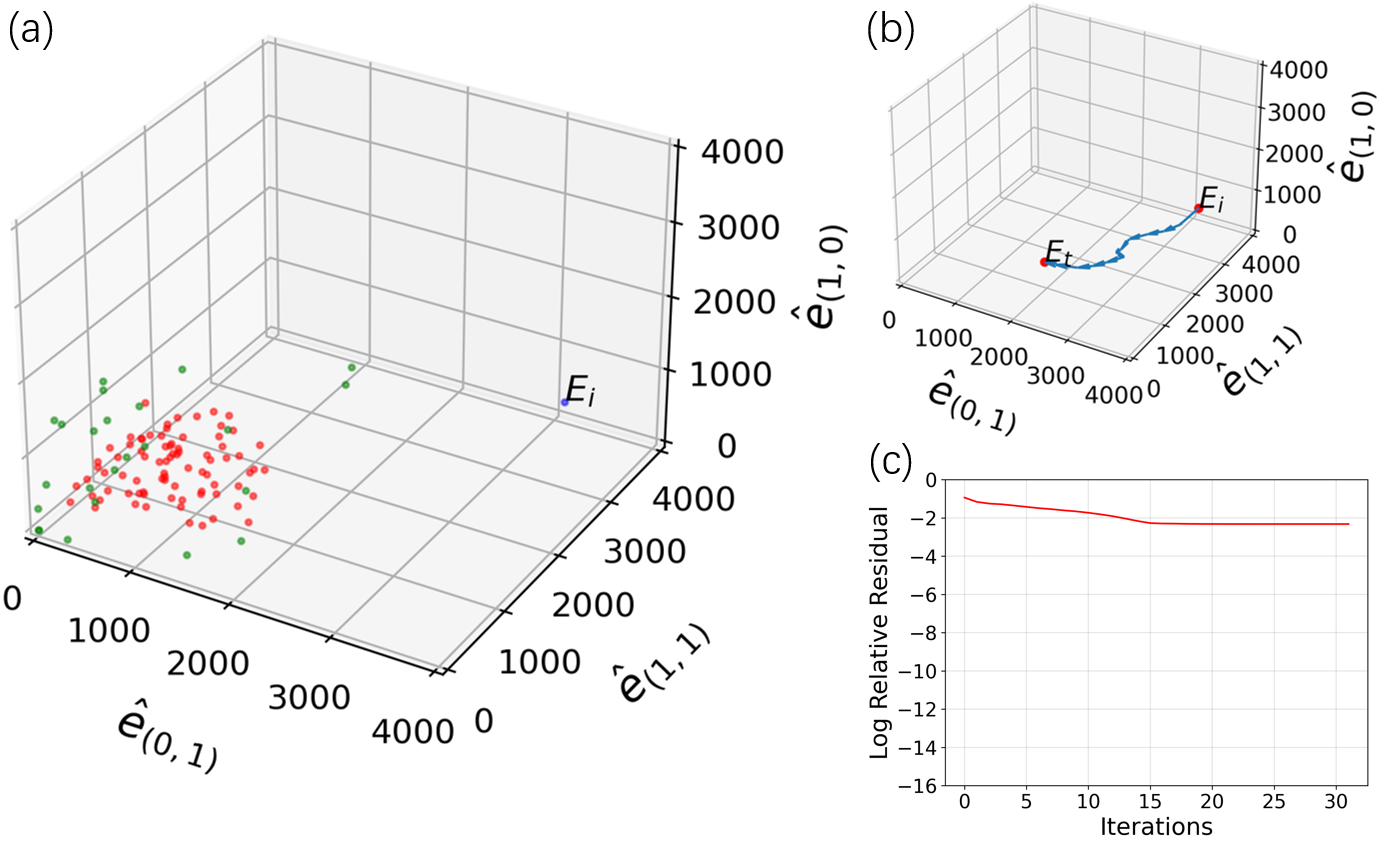}
  \caption{Convergence/divergence of RL+JFNK and JFNK methods. (a) blue point $E_i$: the initial point; red points: the DRL-assisted initial guesses obtained through exploration starting from the initial point by employing the DRL algorithms; green points: fixed points obtained through JFNK method starting from the DRL-based initial guesses. {\color{black}(b) trajectories of the JFNK method from $E_i$ without the assistance of DRL. }(c) log relative residual of the JFNK method. }
    \label{Fig:FindPointsEffect}
\end{figure}

In the following, we provide numerical results of this combined method. As illustrated in Figure \ref{Fig:FindPointsEffect}{\color{red}(a)}, the blue point $E_i$ represents the initial point in the spectral space {$\widehat{e}_{(0,1)}$, $\widehat{e}_{(1,1)}$, $\widehat{e}_{(1,0)}$} (see the next section for a detailed explanation of these symbols. For the moment, it suffices to know them as spectral components) and the red points represent the DRL-based initial guesses obtained through exploration starting from the blue initial point $E_i$ by employing the aforementioned DRL algorithms. The green points signify the fixed points obtained through the JFNK method starting from the DRL-based initial guesses. These are the genuine fixed points in the 2D KSE. {\color{black}It is important to note that the green points} could not be obtained solely using the JFNK method. Indeed, starting from the $E_i$, using only the JFNK method will result in non-convergence, as illustrated in panels (b) and (c) of the figure. Thus, we have demonstrated that DRL can alleviate the impediments in the JFNK method by providing superior initial guesses (red points in panel a) characterized by relatively smaller residuals. When these initial guesses are fed to the JFNK, the latter can produce converged fixed points fast. Therefore, DRL emerges as a viable strategy to address and potentially rectify the issues of non-convergence that stem from insufficient initial guesses in the JFNK method.


\subsubsection{DRL-enhanced initial conditions}

{\color{black} As suggested by one of the reviewers, we explain our methodology for generating and processing the initial guesses, as follows. First, we generate a 64$\times$64 two-dimensional matrix (corresponding to the Fourier collocation points for spatial discretisation), where each entry of the matrix is randomly drawn from a uniform distribution in the range [0, 1). Second, this random initial state is evolved within the KS environment for 1000 time steps to ensure that the initial state is close to the unforced KS manifold, although 500 time steps are typically adequate to bring the state near the unforced KS manifold. The resulting initial conditions are then used in both the classic JFNK algorithm and the DRL-JFNK algorithm developed in this study. All the random initial guesses or conditions mentioned in this work refer to those generated using this method. 

Next, we explain in detail how the initial conditions are improved in the DRL framework. In the DRL training process, the model is trained for 500 episodes or over. We limit each episode to 500 steps to prevent the total training time from becoming excessive. Once this step limit is reached, the episode is terminated. Each episode begins with a random initial guess, where the agent explores the 2D KSE environment to identify a suitable DRL-based initial guesses. During the 500-step exploration, we record the state with the highest reward as a potential DRL initial guess. A reward threshold, denoted as $\theta_{th} = -45$, has been established. If the highest reward of the potential DRL initial guess is greater than $\theta_{th}$, the state is passed to the JFNK method for converging. Otherwise, the initial condition is abandoned.


\begin{figure}
  \centering
  \includegraphics[width=1.0\textwidth]{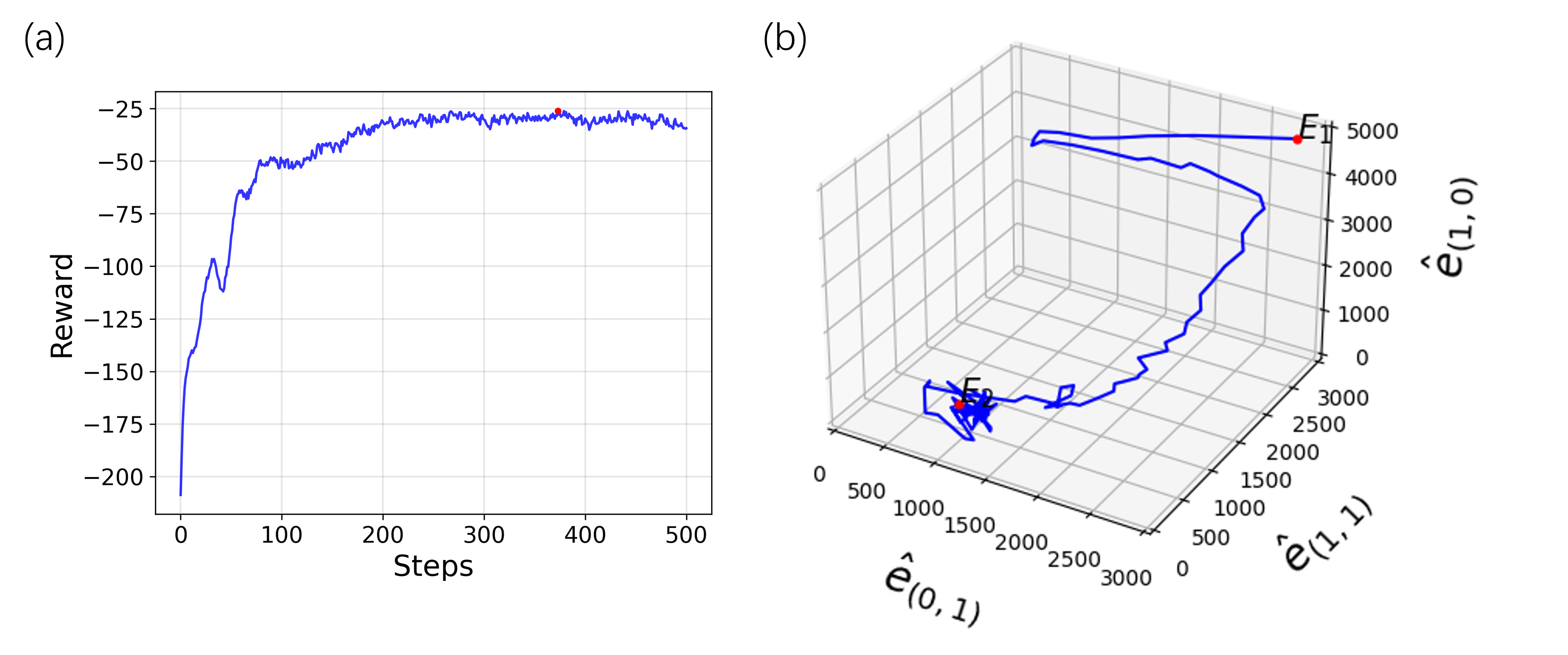}
  \caption{{\color{black} (a) The process for DRL agent to find a potential DRL initial guess. Blue trajectory: the evolution of the DRL agent's exploration of the 2D KSE environment over the course of an episode. Red point: the state with the highest reward found during this exploration. (b) the evolution of the DRL state in the Fourier space (blue trajectory). Red point $E_1$: random initial guess; red point $E_2$: the state with the maximum reward.}}
    \label{Fig:FindPointsEffect2}
\end{figure}

For example, figure \ref{Fig:FindPointsEffect2} presents the process of DRL searching for a successful initial guess. In panel (a), the blue trajectory represents the evolution of the DRL agent's exploration of the 2D KSE environment over the course of an episode, with each point indicating the reward value at a specific time step. The red-marked point denotes the state with the highest reward found during this exploration. Since its reward exceeds the threshold  $\theta_{th} = -45$, this point is selected as a DRL-based initial guess and passed to the JFNK method for further processing. The corresponding evolution of the solution state in the Fourier space is shown in panel (b). 
}

\subsubsection{A comparative test}

\begin{figure}
  \centering
  \includegraphics[width=1.0\textwidth]{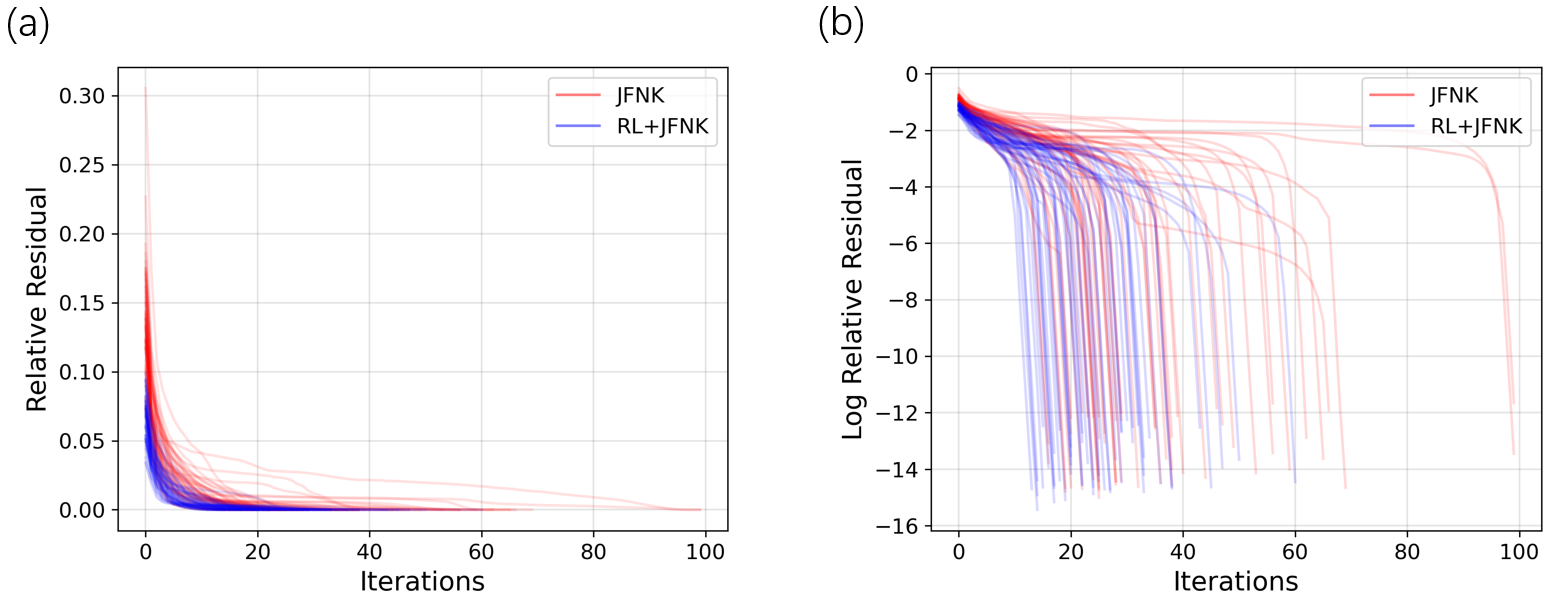}
  \caption{JFNK convergence processes of random initial guesses (red curves) and DRL-enhanced initial guesses (blue curves) in (a) linear stable and (b) log scale. 50 random initial guesses and 50 DRL-based initial guesses were tested, which can converge to the fixed points. }
    \label{Fig:FindPointsCompare}
\end{figure}

The above results demonstrated the effectiveness of adopting the DRL method to provide good initial guesses for JFNK. Next, we prove that the DRL method can also accelerate the convergence process. Based on the same random-number generator, we conducted two experiments: (1) passing random initial guesses to the JFNK method directly; (2) feeding the same initial guesses to the DRL method, the results of which will then be passed to JFNK. We found that the DRL-based initial guesses markedly reduce the number of iterations compared to the sole JFNK method, as shown in Figure \ref{Fig:FindPointsCompare}. The above procedure is applied to generate 50 converged instances starting from the random initial guesses using the JFNK method only and 50 converged instances with the DRL-assisted initial guess fed to the JFNK method.  
Under this comparison, it is evident that the DRL-based initial points exhibit on average smaller relative residuals, leading to fewer required iterations in JFNK. Statistically, the results indicate an average process of 38.2 iterations for convergence with random initial guesses using JFNK solely, in contrast to 27.4 iterations with DRL-based guesses. This reduction in the number of iterations not only underscores the efficiency of DRL-informed initial guesses but also implies a consequent decrease in computational costs.

After the explanation and demonstration of the advantage of the DRL-based initial guesses, we will provide all the fixed points that have been identified in the next section.

\subsection{Fixed points in 2D KSE}\label{fixedpoints}

In this work, we focus exclusively on the most widely studied case where $L_x=L_y$ in the 2D KSE. Kalogirou \textit{et al.} \cite{2DKSE} elucidated that the domain sizes in the 2D KSE exert a strong influence on the system's dynamics. An increment in the domain size causes a transition in the solutions of the equation, evolving from the steady state through a series of changes: periodic homoclinic bursts, periodic heteroclinic bursts, chaotic heteroclinic bursts, travelling waves, and eventually to chaotic solutions. An expansion in the domain size is directly proportional to an increase in the chaotic nature of the 2D KSE. Augmenting in domain sizes also necessitates a corresponding increase in the number of Fourier collocation points, which results in greater computational resource costs. Consequently, we have opted for $2L=2L_x = 2L_y = 20$ as our domain sizes. Although this represents a comparatively lower value, it is sufficient to maintain the chaotic nature of the 2D KSE. We will attempt to employ the JFNK method to explore the fixed points of the 2D KSE. In cases where JFNK did not converge, we implemented the aforementioned DRL approach to generate effective initial guesses. We have successfully identified a total of 303 fixed points, as listed in tables \ref{tab:fixed_points} and \ref{tab:zero_fixed_points}, each representing a distinct and unique solution to the 2D KSE system.


\begin{table}
    \setlength{\arrayrulewidth}{0.3mm}     
    \setlength{\tabcolsep}{1.6pt} 
    \renewcommand{\arraystretch}{1.25}

    \tiny
    \begin{tabular}{c c c c|cc c c|c c c c |c c c c | c c c c} 

      No.& $\widehat{e}_{(0,1)}$& $\widehat{e}_{(1,1)}$&$\widehat{e}_{(1,0)}$& No.&  $\widehat{e}_{(0,1)}$& $\widehat{e}_{(1,1)}$&$\widehat{e}_{(1,0)}$ & No.&  $\widehat{e}_{(0,1)}$& $\widehat{e}_{(1,1)}$&$\widehat{e}_{(1,0)}$ & No.& $\widehat{e}_{(0,1)}$& $\widehat{e}_{(1,1)}$&$\widehat{e}_{(1,0)}$ 
 & No.&  $\widehat{e}_{(0,1)}$& $\widehat{e}_{(1,1)}$&$\widehat{e}_{(1,0)}$
\\
      \hline
      E1$^\dagger$& 0& 0& 0
& E71
& 9& 805.1& 9
& E141
& 469.9& 798.6&757.4
&  
E211
& 1193.5& 1376.6&2020.6
&  
 
E281
& 2526.6& 80.1&1318.1
\\
      E2$^\dagger$& 0& 0& 0
& E72
& 
10.8& 1682.7& 516.1
&  
E142
& 472.9& 180.2&718.5
& E212
& 1226.9& 1468.6& 0
& E282
&  
2556.5& 0&0
\\
 E3$^\dagger$& 0& 0& 0
& E73
& 21.8& 1995.8& 21.8
&  E143
& 474.4& 1624.8&868.6
& E213
& 1237& 778.3&1237
&

E283
&  
2557.7& 1415.7&977.2
\\
 E4$^\dagger$& 0& 0& 0
& E74
& 
22.9& 19& 616.8
&  E144
& 478.6& 178.1&533
&  E214
& 1249.8& 528.9&2246.3
&  
 
E284
&   2563.4& 677.5&2563.4
\\
 E5$^\dagger$& 0& 0& 0
& E75
& 26.2& 446.7& 1631.8
& E145
& 482.1& 294.5&482.1
& E215
& 1301.2& 367&364
& E285
&   2655.3& 871.2&2655.3
\\
 E6$^\dagger$& 0& 0& 0
& E76
& 
33.3& 15.8& 662.1
& E146
& 486.9& 1575.4&486.9
& E216
& 1316.5& 398.9&1316.5
&

E286
& 2782.8& 1256.7&1942.9
\\
 E7$^\dagger$& 0& 0& 0
& E77
& 54.6& 763.3& 142.4
& E147
& 494.9& 178.8&1428.2
& E217
& 1317.9& 2033&1317.9
&

 E287
& 2857.3& 249.3&0
\\
 E8$^\dagger$& 0& 0& 0
& E78
& 
56.2& 1571.5& 31.3
& E148
& 505.1& 202.2&292.7
& E218
& 1318.5& 627.6&1646.9
& E288
& 2930.6& 292.9&2930.6
\\
 E9$^\dagger$& 0& 0& 0
& E79
& 56.6& 1737.6& 56.6
& E149
& 512& 496.4&597
& E219
& 1324& 270.4&302.6
&

  E289
& 3114.6& 689.3&1420.6
\\
 E10$^\dagger$& 0& 0& 0
& E80
& 
65.7& 164.4& 1899.9
& E150
& 524.7& 3234.6&524.7
& E220
& 1335.3& 423.5&137.4
&  
 
  E290
& 3192.5& 1661.2&2598.3
\\
 E11$^\dagger$& 0& 0& 0
& E81
& 68.6& 122.6& 718
& E151
& 527.5& 114&527.5
& E221
& 1345.4& 393.2&424.1
& E291
& 3351.2& 352.1&2071.9
\\
 E12$^\dagger$& 0& 0& 0
& E82
& 
74.2& 874.6& 783.3
& E152
& 528.3& 2106.5&6613.1
& E222
& 1345.5& 0&0
&

E292
& 3731.1& 2573.9&5433.7
\\
 E13$^\dagger$& 0& 0& 0
& E83
& 76.6& 1989.5& 56.6
& E153
& 535.5& 771.4&2042.6
& E223
& 1352.1& 58.3&1352.1
&  
 
E293
& 3811.3& 718&5056.4
\\
 E14$^\dagger$& 0& 0& 0
& E84
& 
77.5& 1095& 77.5
& E154
& 544.7& 1003.6&1016.7
& E224
& 1355& 1165&1355
& E294
& 4056.5& 2853.2&5817.2
\\
 E15$^\dagger$& 0& 0& 0
& E85
& 92.2& 105.4& 92.2
& E155
& 554& 2808.7&2187.5
& E225
& 1358.1& 873.4&1488.1
&

E295
& 4754.1& 700.8&4754.1
\\
 E16$^\dagger$& 0& 0& 0
& E86
& 
94.5& 6.1& 668.6
& E156
& 567.9& 1139.8&2512.6
& E226
& 1362.9& 92.7&1125.5
&  
 
E296
& 5100& 3826.3&5100
\\
 E17$\dagger$
& 0& 0& 0
& E87
& 97.9& 2847.2& 1046.6
& E157
& 576.1& 257.3&1513.7
& E227
& 1379.3& 178.9&1675.6
& E297
& 5320.8& 4279&2242.6
\\
 E18
& 0& 0& 129.7
& E88
& 
100& 1809.9& 1660.2
& E158
& 577.3& 324.9&1275.5
& E228
& 1395& 1031.6&1395
&

E298
& 5464.5& 1546.4&5464.5
\\
 E19
& 0& 0& 302
& E89
& 116.5& 845.8& 14.1
& E159
& 585.1& 352.5&634.5
& E229
& 1397.9& 416.4&0
&  
 
E299
& 5583.1& 6174.4&5583.1
\\
 E20
& 0& 0& 852.5
& E90
& 
123.4& 670.1& 445.7
& E160
& 590.1& 761.9&1400
& E230
& 1455.6& 396.6&1455.6
& E300
& 6096.7& 498.4&0
\\
 E21
& 0& 0& 1246.8
& E91
& 130.7& 400.1& 421.5
& E161
& 590.7& 137&122.9
& E231
& 1459.7& 1816.4&1459.7
&  
   E301
& 6331.2& 0&6331.2
\\
 E22
& 0& 0& 1295.5
& E92
& 

134.8& 809& 281.5
& E162
& 607.8& 189.9&607.8
& E232
& 1460.6& 2791.6&2792.1
&  
 
E302
& 6637.3& 5377.4&6637.3
\\
 E23
& 0& 0& 2000.8
& E93
& 142.4& 237.8& 2454.2
& E163
& 609.5& 300.6&1361.2
& E233
& 1460.9& 346.1&1460.9
& E303& 6744.6& 922&1570.2
\\
 E24
& 0& 0& 2059.3
& E94
& 

149.4& 212.9& 2574.3
& E164
& 611& 1021.7&437.3
& E234
& 1470& 231.5&664.4
&  
& & &\\
 E25
& 0& 0& 2553.9
& E95
& 151.7& 186& 313.3
& E165
& 616.1& 211.4&1619.8
& E235
& 1485.4& 921.9&1485.4
&  
 
& & &\\
 E26
& 0& 0& 2871.6
& E96
& 

161.3& 768.2& 161.3
& E166
& 624.8& 1534.9&906.2
& E236
& 1498.9& 131.5&824.8
& & & &\\
 E27
& 0& 0& 6331.2
& E97
& 175& 2395.5& 258.5
& E167
& 625.5& 1786.8&1206.1
& E237
& 1501.7& 765.1&1197.3
&  
& & &\\
 E28
& 0& 0& 6331.2
& E98
& 

181.7& 1526.8& 1382.9
& E168
& 627.2& 1050.4&860.9
& E238
& 1502.4& 1342.2&1502.4
&  
 
& & &\\
 E29
& 0& 59.2& 2391.3
& E99
& 182.4& 1872.1& 385.5
& E169
& 630.2& 44.1&1791.1
& E239
& 1507.2& 1089.5&199.2
& & & &\\
 E30
& 0& 64.2& 867.9
& E100
& 

182.8& 283& 113.9
& E170
& 635.9& 117.4&1202.4
& E240
& 1507.5& 100.4&1507.5
&  
& & &\\
 E31
& 0& 138.3& 0
& E101
& 193.9& 32.9& 1551.8
& E171
& 637.9& 1033.8&1689.7
& E241
& 1519.5& 451.8&1519.5
&  
 
& & &\\
 E32
& 0& 222.8& 2386.5
& E102
& 

195.7& 398.6& 137.9
& E172
& 655& 599.5&335.3
& E242
& 1535.8& 596.3&1535.8
& & & &\\
 E33
& 0& 229.5& 1154.8
& E103
& 207.1& 1692.7& 188.1
& E173
& 660.9& 903.2&3276.7
& E243
& 1547.2& 742.1&534.9
&  
& & &\\
 E34
& 0& 404.9& 1549.4
& E104
& 

217.4& 770.2& 544
& E174
& 694.2& 198.7&694.2
& E244
& 1554& 84.2&722.6
&  
 
& & &\\
 E35
& 0& 419.3& 0
& E105
& 219.5& 102.3& 1465.9
& E175
& 705.4& 965.4& 929.1
& E245
& 1570.5& 630.7&1570.5
& & & &\\
 E36
& 0& 485.6& 0
& E106
& 224.6& 413.4& 224.6
& E176
& 708.1& 41.5& 708.1
& E246
& 1579.4& 303.5& 1049.1
& & & &\\
 E37
& 0& 489.2& 0
& E107
& 234.2& 1024.7& 649.7
& E177
& 718.9& 1429.3& 195.4
& E247
& 1607.6& 579.3& 200
& & & &\\
 E38
& 0& 495.2& 1654.9
& E108
& 236.1& 155.9& 319.8
& E178
& 727.1& 610.3& 2548.9
& E248
& 1628.9& 0& 0
& & & &\\
 E39
& 0& 516.7& 0
& E109
& 250& 241.8& 92.4
& E179
& 735.1& 1025.9& 842.8
& E249
& 1634.5& 1153.9& 1634.5
& & & &\\
 E40
& 0& 551.6& 0
& E110
& 252.6& 123.7& 1364.6
& E180
& 742.9& 44.4& 229.5
& E250
& 1639.7& 1140.1& 1639.7
& & & &\\
 E41
& 0& 649.8& 0
& E111
& 262.9& 1279.5& 100.1
& E181
& 752.7& 340.3& 2515.6
& E251
& 1657& 2634.8& 2128
& & & &\\
 E42
& 0& 747& 0
& E112
& 264.2& 80.5& 1251
& E182
& 773.5& 619.4& 1560.4
& E252
& 1715.2& 66.5& 1715.2
& & & &\\
 E43
& 0& 766.5& 0
& E113
& 266.8& 1575.8& 290.3
& E183
& 798.1& 939& 700.5
& E253
& 1751.4& 1005.3& 1655.4
& & & &\\
  E44
& 0& 790.8& 2725.3
& E114
& 289.9& 1383.5& 241.6
& E184
& 803.6& 73.5& 1785.2
& E254
& 1772& 361.4& 1772
& & & &\\
 
E45
& 0& 797& 0
& E115
& 294.1& 678.6& 0
& E185
& 806.1& 2934& 806.1
& E255
& 1799.6& 2019.6& 1799.6
& & & &\\
 E46
& 0& 811.2& 0
& E116
& 

295.5& 1260.5& 295.5
& E186
& 838.6& 1098& 838.6
& E256
& 1810.1& 133.6&262.4
&  
& & &\\
 E47
& 0& 817.6& 0
& E117
& 295.9& 329& 295.9
& E187
& 843.1& 1109.5& 843.1
& E257
& 1825.8& 45.1&558.6
&  
 
& & &\\
 E48
& 0& 905.3& 1051.2
& E118
& 

313.8& 65.3& 750.6
& E188
& 846.7& 570.9& 1254.6
& E258
& 1844.3& 1567.5&1730.9
& & & &\\
 E49
& 0& 1053.1& 0
& E119
& 323.1& 465.7& 538.8
& E189
& 860.5& 2895.9& 860.5
& E259
& 1889.6& 1431.1&2456.8
&  
& & &\\
 E50
& 0& 1160.5& 0
& E120
& 

324.3& 488.3& 804.5
& E190
& 863& 610.7& 863
& E260
& 1913.5& 2285.3&1913.5
&  
 
& & &\\
 E51
& 0& 1227.6& 0
& E121
& 333.4& 620.1& 1056.2
& E191
& 879.8& 1106& 719.5
& E261
& 1920.6& 252.4&496.7
& & & &\\
 E52
& 0& 1238.1& 419.2
& E122
& 

334.5& 810.6& 334.5
& E192
& 883.1& 1537.4& 39.2
& E262
& 1932.5& 2086.4&1493.7
&  
& & &\\
 E53
& 0& 1251.2& 0
& E123
& 348.3& 748.3& 1930
& E193
& 887.7& 1420.7& 2602.3
& E263
& 1970.8& 760.6&2302.2
&  
 
& & &\\
  E54
& 0& 1311.1& 0
& E124
& 

356.3& 26.6& 431.5
& E194
& 916.1& 1233.9& 1676
& E264
& 1974.8& 26.6&3413.2
& & & &\\
 
E55
& 0& 1343.6& 0
& E125
& 360.6& 849.6& 1262.2
& E195
& 958.9& 1924& 1812.4
& E265
& 1980.2& 375&3789.9
&  
& & &\\
 E56
& 0& 1370.4& 852
& E126
& 

364& 128.1& 364
& E196
& 961.3& 1168.3& 961.3
& E266
& 1982.2& 0&0
&  
 
& & &\\
 E57
& 0& 1411.9& 0
& E127
& 366.5& 324.8& 1924.1
& E197
& 971.3& 1106.7& 1062
& E267
& 2031.1& 2082.9&2824.6
& & & &\\
 E58
& 0& 1428.2& 0
& E128
& 

381.4& 1749.7& 555.5
& E198
& 979.4& 166.5& 979.4
& E268
& 2096.6& 2584.5&2096.6
&  
& & &\\
 E59
& 0& 1454.7& 0
& E129
& 385& 1135.6& 456.3
& E199
& 1001.9& 949.6& 427.8
& E269
& 2132.7& 1491.9&2436.2
&  
 
& & &\\
 E60
& 0& 1474.6& 552.5
& E130
& 

392& 853.2& 1258.3
& E200
& 1002& 690.9& 1532
& E270
& 2205.6& 3864.8&2205.6
& & & &\\
 E61
& 0& 1502.4& 0
& E131
& 393.7& 479.3& 141.6
& E201
& 1018.4& 310& 2070.3
& E271
& 2236.7& 1817.4& 2236.7
& & & &\\
 E62
& 0& 1524.6& 2321.1
& E132
& 409& 663.9& 951.7
& E202
& 1054.3& 606.8& 1340.4
& E272
& 2251.5& 399.6& 1565.7
& & & &\\
 E63
& 0& 1642.2& 44.8
& E133
& 409.9& 3224.2& 409.9
& E203
& 1056.8& 346.7& 300.3
& E273
& 2257.7& 2709.7& 2257.7
& & & &\\
 E64
& 0& 1681.2& 0
& E134
& 415.1& 374.6& 696.5
& E204
& 1110.4& 174& 1311.4
& E274
& 2304.4& 2995.4& 1235.8
& & & &\\
 E65
& 0& 1943& 746.2
& E135
& 438.3& 1628.9& 562.7
& E205
& 1114.4& 1038.3& 520.3
& E275
& 2331.2& 1044.4& 516.3
& & & &\\
 E66
& 0& 2335.6& 536.2
& E136
& 443.6& 1052.3& 176.4
& E206
& 1117.5& 1220.5& 1117.5
& E276
& 2332.8& 1385.1& 1434.2
& & & &\\
 E67
& 0& 2378.1& 566.2
& E137
& 444.2& 607.7& 603.2
& E207
& 1124.7& 739.1& 1124.7
& E277
& 2399.8& 2037.3& 1505.4
& & & &\\
 E68
& 0& 2868.7& 0
& E138
& 447.6& 2868.6& 447.6
& E208
& 1127.4& 1426.5& 2127.9
& E278
& 2461.7& 1599.9& 937.2
& & & &\\
 E69
& 0.4& 727.3& 0.4
& E139
& 450.7& 3625.9& 450.7
& E209
& 1140.4& 34& 1848.9
& E279
& 2489& 1362.4& 0
& & & &\\
 E70
& 2.8& 271.8& 2223.2
& E140
& 456.7& 2965.5& 456.7
& E210
& 1153.9& 1220& 434.3
& E280
& 2520& 2232.4& 2772.5
& & & &\\
      
    \end{tabular}

  \caption{Fixed points that have been obtained with the aid of DRL in the JFNK method. For conciseness, we tabulate only the absolute value of the first three complex-valued Fourier coefficients {$\widehat{e}_{(0,1)}$, $\widehat{e}_{(1,1)}$, $\widehat{e}_{(1,0)}$} for each point. The absolute value has not been normalised by the spatial dimensions (64$\times$64). Only one decimal place is retained for clarity. \\
  $^\dagger$The cases $E1$-$E17$ are further listed in table \ref{tab:zero_fixed_points}. }
  
  \label{tab:fixed_points}
\end{table}

\begin{table}
    \setlength{\arrayrulewidth}{0.3mm}     
    \setlength{\tabcolsep}{3.2pt} 
    \renewcommand{\arraystretch}{1.25}

    \scriptsize  
    \begin{tabular}{c c c c c c c c | c c c c c c c c} 

      No.& $\widehat{e}_{(2,0)}$& $\widehat{e}_{(2,1)}$& $\widehat{e}_{(3,0)}$&$\widehat{e}_{(3,1)}$& $\widehat{e}_{(0,2)}$ & $\widehat{e}_{(1,2)}$& $\widehat{e}_{(2,2)}$ & No.&  $\widehat{e}_{(2,0)}$& $\widehat{e}_{(2,1)}$& $\widehat{e}_{(3,0)}$&$\widehat{e}_{(3,1)}$& $\widehat{e}_{(0,2)}$ & $\widehat{e}_{(1,2)}$& $\widehat{e}_{(2,2)}$ \\
      \hline
      E1& 2746.8& 0& 0& 0& 0& 4048.3& 0
& E11
& 0& 540.1& 0& 0& 0& 0&0
\\
      E2& 2367& 0& 0& 0& 0& 1866.4& 0
& E12
& 0& 0& 0& 1606.5& 0& 0&0
\\
 E3& 0& 5914& 0& 0& 0& 0& 0
& E13
& 2175.2& 0& 0& 0& 2175.2& 0&901.2
\\
 E4& 4171.3& 1925.4& 0& 0& 4143.8& 0& 308.2
& E14
& 0& 0& 0& 0& 5086.6& 0&0
\\
 E5& 0& 5543.4& 0& 0& 0& 0& 0
& E15
& 0& 0& 4387.2& 0& 0& 0&0
\\
 E6& 0& 0& 0& 0& 0& 5914& 0
& E16
& 5086.6& 0& 0& 0& 5086.6& 0&0
\\
 E7& 0& 0& 0& 0& 0& 0& 1293
& E17& 0& 0& 0& 993.6& 0& 0&2201.8
\\
 E8& 0& 0& 0& 1394.6&0& 0& 318.8
& & & & & & & &\\
 E9& 0& 0& 0& 0&0& 0& 0
& & & & & & & &\\
 E10& 0& 788.4& 1870& 0&0& 788.4& 0
& & & & & & & &\\
      
    \end{tabular}

  \caption{Representation of the fixed points E1-E17 in 2D KSE which all have the same $\widehat{e}_{(0,1)}=\widehat{e}_{(1,1)}=\widehat{e}_{(1,0)}=0$. }
  
  \label{tab:zero_fixed_points}
\end{table}

\begin{figure}
  \centering
  \includegraphics[width=0.65\textwidth]{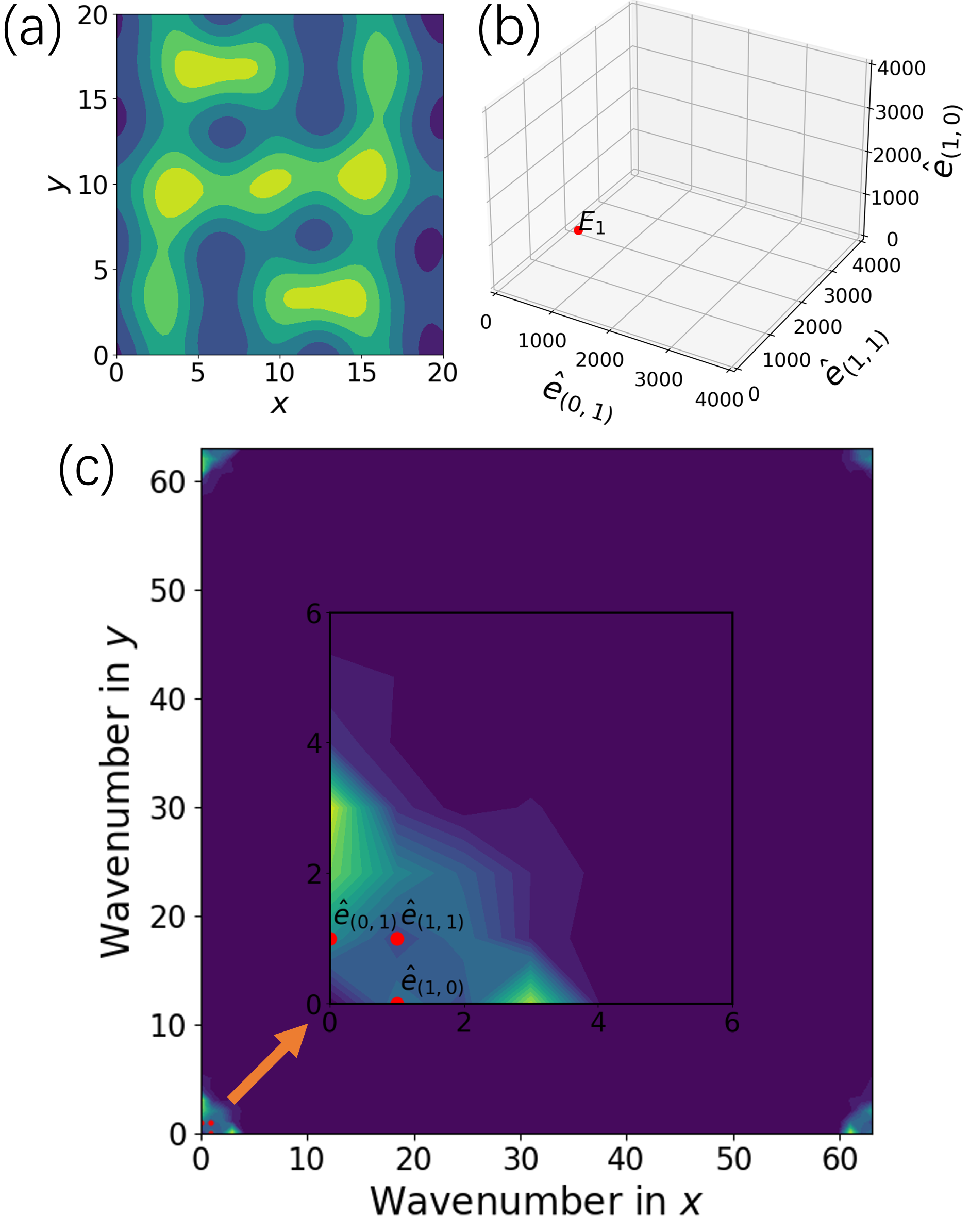}
\caption{Examples of fixed points in the 2D KSE: (a) spatial distribution and (b) in phase space. Panel (c) shows the three modes in the Fourier space. }
    \label{Fig:FixedPoint}
\end{figure}

In order to illustrate the fixed point as the solution to the 2D KSE, as shown in Figure \ref{Fig:FixedPoint}, we utilize the three representations to depict the fixed point: spatial domain, phases space, and Fourier space. The Fourier space is derived through a Fast Fourier Transformation (FFT) in space, yielding a 64$\times$64 matrix that mirrors the dimensions of its discretised spatial matrix. In the phase space representation, the figure is characterized by the set {$\widehat{e}_{(0,1)}$, $\widehat{e}_{(1,1)}$, $\widehat{e}_{(1,0)}$} in the Fourier space. $\widehat{e}_{(0,1)}$, $\widehat{e}_{(1,1)}$, $\widehat{e}_{(1,0)}$ are the absolute values of the 2D FFT coefficients without normalisation by the spatial dimensions (they are the direct result of the fft2 command in Matlab). More specifically, the lower-left point of the Fourier space in figure \ref{Fig:FixedPoint}(c), i.e., the zero-wavenumber component in $x$ and $y$ directions, represents the mean mode. The three points close to the mean mode are defined as the {$\widehat{e}_{(0,1)}$, $\widehat{e}_{(1,1)}$, $\widehat{e}_{(1,0)}$} modes for simplicity. $\widehat{e}_{(0,1)}$ is the first Fourier mode in the $y$ direction with zero Fourier mode in the $x$ direction; $\widehat{e}_{(1,0)}$ denotes the first Fourier mode in the $x$ direction with zero Fourier mode in the $y$ direction; and $\widehat{e}_{(1,1)}$ represents the first Fourier mode in both $x,y$ directions, see the three red points in the lower-left corner of the Fourier space in Figure \ref{Fig:FixedPoint}. All the identified fixed points are listed in table \ref{tab:fixed_points} with these three Fourier components. Note that because we only used three modes to represent the high-dimensional numerical results in the 2D KSE, in some cases the values of $\widehat{e}_{(0,1)}$, $\widehat{e}_{(1,1)}$, $\widehat{e}_{(1,0)}$ appear the same for distinct fixed points. Under these circumstances, more Fourier coefficients will be provided to differentiate them (see table \ref{tab:zero_fixed_points}).

\begin{figure}
  \centering
  \includegraphics[width=1\textwidth]{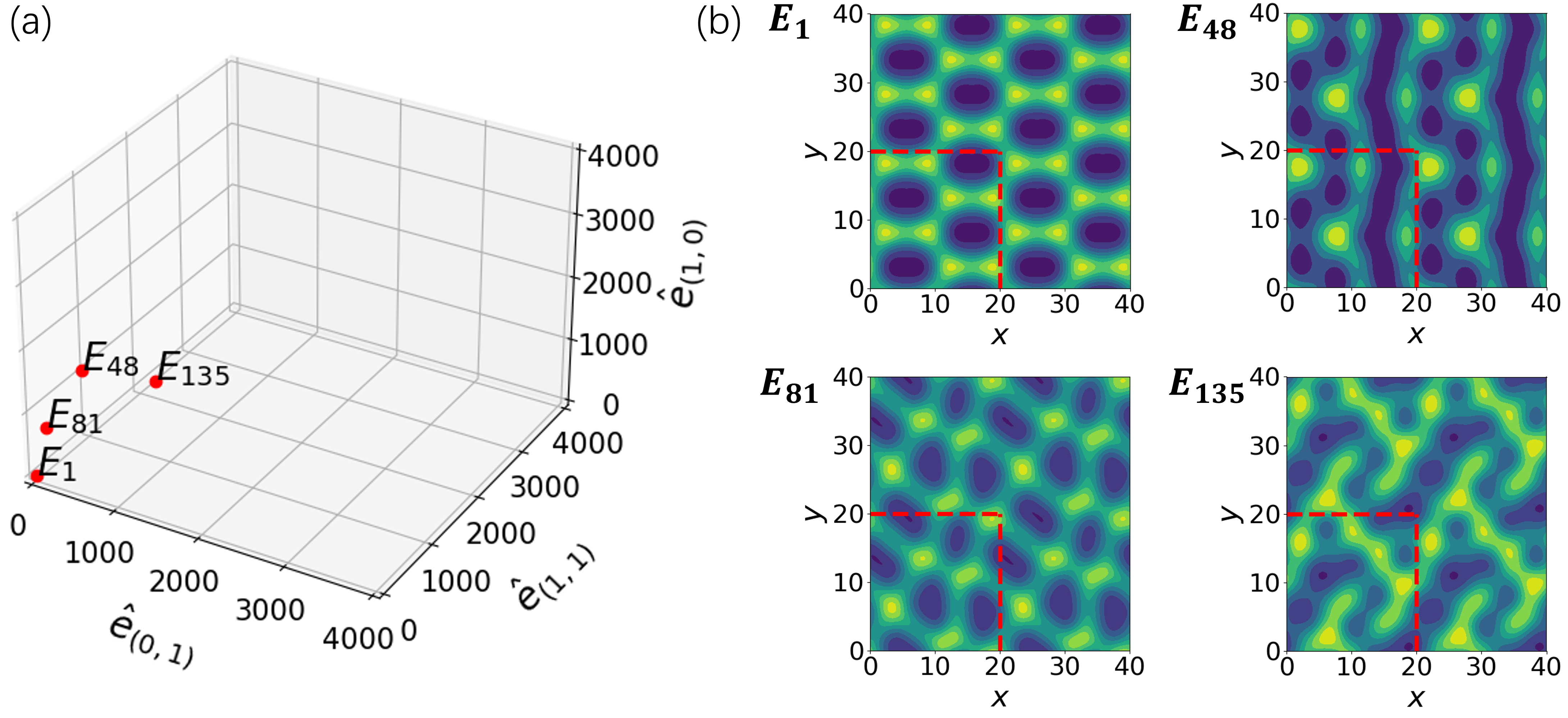}
  \caption{Four exemplary fixed points in the 2D KSE in (a) phase space and (b) physical spatial domain. The spatial distributions of the fixed points $E_1$ and $E_{48}$ exhibit symmetric periodic behavior, while $E_{81}$ and $E_{135}$ exhibit non-symmetric periodic behaviour. These four points are also presented in Table \ref{tab:fixed_points}. The numerical identifier assigned to each point corresponds to its sequential order as listed in the table.}
    \label{Fig:FixedPoints}
\end{figure}

Due to space constraints, we have selected to illustrate four representative fixed points as examples in Figure \ref{Fig:FixedPoints}. The four fixed points are displayed in the phase space in panel (a). Their spatial representations are shown in panel (b). Although our computations were conducted with $2L=2L_x=2L_y=20$ (see the red dashed box in the panel), in order to visually demonstrate the periodicity of the solution, we extend the results to $[0,40]\times[0,40]$. As shown in panel (b), the fixed point $E_1$ exhibits pronounced symmetry in both the $x,y$ directions. Similarly, $E_{48}$ demonstrates symmetry along the $y$ direction. Meanwhile, $E_{81}$ and $E_{135}$ do not present any evident symmetrical properties in any directional axis. From these four fixed points, it is evident that certain fixed points exhibit symmetries, while others display asymmetry, reflecting the diversity and complexity of the 2D KSE system. 

\begin{figure}
  \centering
  \includegraphics[width=0.4\textwidth]{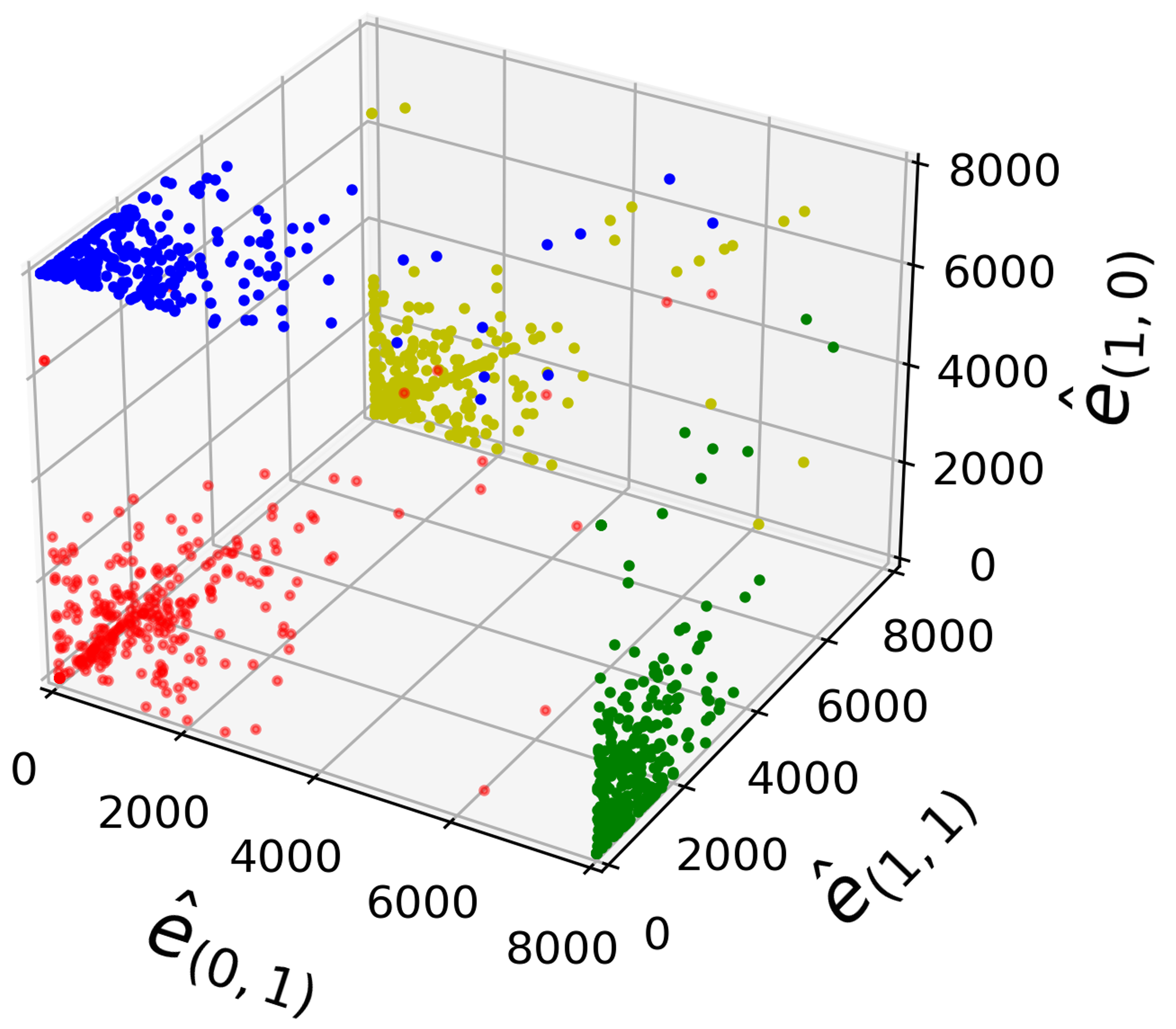}
  \caption{Fixed points (red) in the phase space of the 2D KSE. To better visualize the fixed points, we also plot the projection of the red dots on the $(\widehat{e}_{(0,1)},\widehat{e}_{(1,1)})$, $(\widehat{e}_{(0,1)},\widehat{e}_{(1,0)})$, $(\widehat{e}_{(1,1)},\widehat{e}_{(1,0)})$ planes, see the blue, yellow, and green dots. Note that the values of $\widehat{e}_{(0,1)},\widehat{e}_{(1,1)}, \widehat{e}_{(1,0)}$ have not been normalised against the spatial dimensions. }
      \label{Fig:all_points}
\end{figure}

In order to gain an overview of the distribution of all the identified fixed point in the phase space, we plot them  in figure \ref{Fig:all_points} as the red points. The projections of the fixed points on the $(\widehat{e}_{(0,1)},\widehat{e}_{(1,1)})$, $(\widehat{e}_{(0,1)},\widehat{e}_{(1,0)})$, $(\widehat{e}_{(1,1)},\widehat{e}_{(1,0)})$ planes are denoted by the blue, yellow, green points, respectively. One can see that the distribution of fixed points is dense in the corner close to the original point. As the values of {$\widehat{e}_{(0,1)}$, $\widehat{e}_{(1,1)}$, $\widehat{e}_{(1,0)}$}  increase further, the distribution of fixed points becomes sparser. 

\subsection{DRL-based navigation between fixed points}

\begin{figure}[!h]
  \centering
  \includegraphics[width=0.9\textwidth]{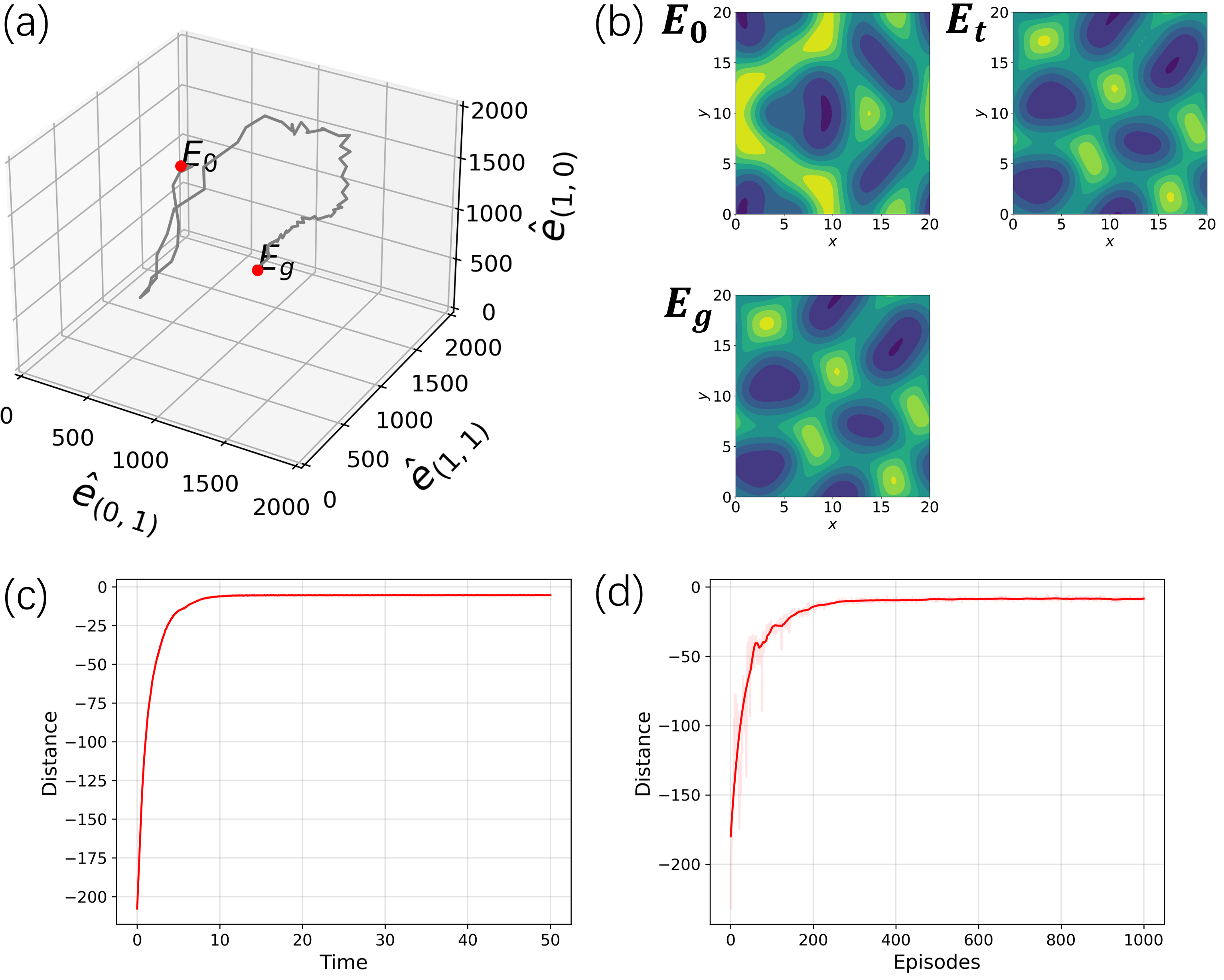}
  \caption{DRL-based navigation between fixed points in 2D KSE and the DRL training process. Panel (a) shows the evolution of a trajectory (grey curve) emanating from the initial state $E_0$ to the goal state $E_g$ in the phase space. Panel (b) presents the spatial profile of the state $E_0$ at $t=0$, $E_t$ at $t = 50$ and the goal state $E_g$. $E_t$ is the state in the evolution, which is visually the same as $E_g$. Panels (c) and (d) are the negative distance between the agent's state and the goal state in the test case and the DRL training process as a function of time. In (d), we collect the distance at the conclusion of each episode.}
    \label{Fig:Control_Figure}
\end{figure}
Next, we illustrate the ability of DRL agent of navigation to the goal fixed point {\color{black}in the 2D KSE phase space}. In alignment with the principles of DRL, the agent is programmed to persistently increase its reward. As delineated in Section \ref{method}\ref{reward_section}, the reward of this task is defined as the negative of the Euclidean distance between the current state of the system and the goal fixed point, which is known. Consequently, the agent's objective is to approximate and subsequently maintain proximity to this goal fixed point of the 2D KSE.

However, the agent's convergence can be impeded by various challenges, such as complex or unstable environments, suboptimal hyperparameter setting, imbalance in exploration and exploitation. This issue also manifested in our research, particularly when directly applying the open-sourced code from Bucci \textit{et al.} \cite{chaotic}, which was originally designed for 1D KSE. To address these impediments, a series of modifications were undertaken, as elaborated in Appendix B\ref{parallel_section} and \ref{noise_section}.

Now we will explain the result of the navigation task to demonstrate that the DRL approach can steer and stabilize the dynamics of the 2D KSE around its unstable fixed solutions. 
As shown in Figure \ref{Fig:Control_Figure}, the navigation is started with a random state $E_0$ to the unstable goal fixed point $E_g$. The gray curve in panel (a) shows the trajectory of the navigation. The shapes of the initial condition $E_0$ at $t=0$, the state $E_t$ at $t=50$ and the goal state $E_g$ are shown in panel (b). As we can see, the state at $t = 50$ closely aligns with the goal state, signalling that the navigation is successful. This congruence is also evident from the distance result in our test case (see panel c), where the DRL agent gets close to the goal state within approximately 8 time units and subsequently maintains stability near this goal point. Besides, from the data of the distance in DRL training process (panel d), it exhibits the capability to converge effectively within as few as 200 episodes, subsequently maintaining stability, demonstrating the training effectiveness and robustness of DRL.

\section{Conclusion}

In this work, we considered employing the JFNK method to identify certain fixed points within the 2D KSE. As well-known, the JFNK approach entails a good initial guess for improved convergence, we proceeded to introduce DRL as a preliminary preprocessing step to address the limitation. {\color{black} The core concept is to leverage the DRL control agent to identify promising initial conditions that exhibit minimal deviation during time evolution. This is guided by the reward function defined within the DRL framework.} This incorporation of DRL into the JFNK framework serves the purpose of facilitating the discovery of enhanced initial guesses. The newly reported more than 300 fixed-points in the 2D KSE demonstrated the effectiveness of the proposed method. These results may also serve as a stepping stone for future investigations of the complex dynamics in the 2D KSE.

{\color{black} Additionally, this work explored several methods to optimize the control of the 2D KSE, including the use of multi-environment reinforcement learning techniques.} This allowed us to navigate the trajectory of the system from one known fixed point to another. Moreover, we undertook the refinement of exploration noise in the DRL framework and utilised parallel agents for the training and testing.

In summation, this work introduced a hybrid approach to enhance the conventional Newton method. {\color{black}The approach began with the application of the JFNK method, followed by the use of DRL as a preprocessing step to improve the efficiency of JFNK.} {\color{black}In a broader context, this work aligns with more advanced applications of general control methods used to
solve fixed-point solutions in chaotic systems  \cite{zeng2021,chaotic,Willis2017}. This work opens several avenues for future research to deepen our understanding of the dynamical systems.} Potential future efforts may adopt the hybrid DRL-JFNK method to solve for the fixed-point or equilibrium solutions in chaotic turbulent flows, {\color{black}such as Couette flow. It would be worthwhile to extend the method to explore other exact coherent structures \cite{Kawahara2011,Graham2021}, such as periodic orbit solutions, in turbulent flows. Additionally, efforts could focus on enhancing the computational efficiency of the combined DRL-JFNK method, including implementing more effective parallelisation, optimising the reward function, and leveraging hardware acceleration. }
\\
\\
Acknowledgment. MZ acknowledges the financial support of a Tier 1 grant A-8001172-00-00 from the Ministry of Education, Singapore. The financial support of NUS (Suzhou) Research Institute and National Natural Science Foundation of China (grant no. 12202300) is also acknowledged. DW is supported by a PhD scholarship (No. 201906220200) from the China Scholarship Council and an NUS research scholarship.

\appendix

\section*{Appendix A: Limitation of JFNK method}

Based on our 1500 test cases, the JFNK method exhibits a significant number of non-convergence instances. In successful cases, the JFNK method has exhibited the capability to reduce the tolerance to below $10^{-10}$, approaching zero with exceptionally high precision, resulting in accurate approximations of the fixed points. However, in instances of failure, the residuals of the JFNK method tend to a plateau at the order between 0.1 to 1, impeding further reduction and leading to convergence failure. Based on the existing 1500 attempts, the probability of failure is approximately around 2/3.

Another issue with the JFNK method is that, under random initial guesses, it may not converge to the fixed point closest in phase space distance. Indeed, the convergence trajectory of the JFNK approach within the phase space can be intricate, often culminating in pronounced discrepancies between the initial and converged points. This complexity frequently results in the JFNK method necessitating additional iterations and, consequently, extended computational time. A comprehensive illustration of this constraint inherent to the JFNK method is provided below.

\begin{figure}
  \centering
  \includegraphics[width=1.0\textwidth]{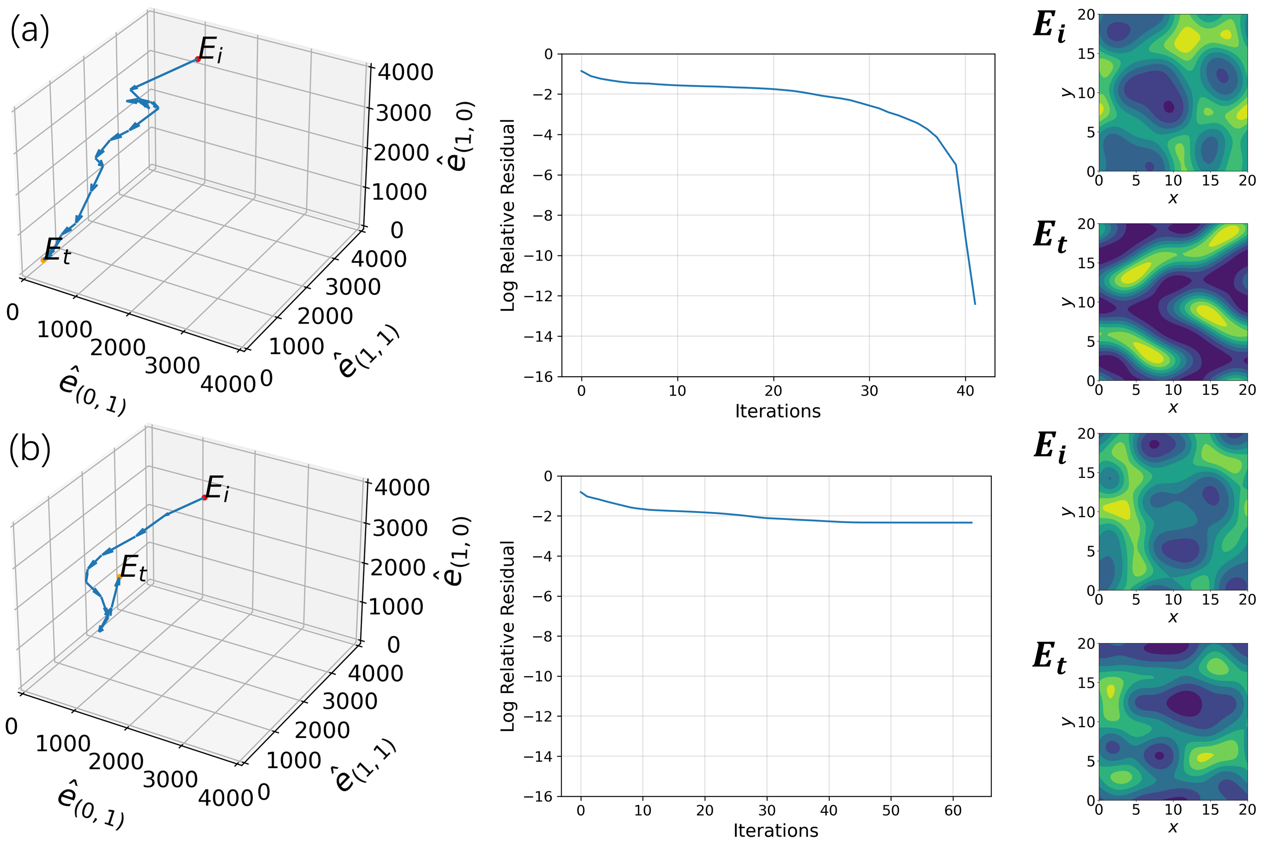}
  
  \caption{Illustration of the JFNK method's convergence process. Panel (a) exemplifies a scenario of successful convergence, while panel (b) illustrates a scenario where convergence is not achieved. The left column depicts the phase space trajectories, characterized by the dominant Fourier coefficients {$\widehat{e}_{(0,1)}$, $\widehat{e}_{(1,1)}$, $\widehat{e}_{(1,0)}$}. The central column presents the log relative residual, capturing the evolution of the JFNK method's residual. The right column shows the initial guess $E_i$ as well as the resultant converged point $E_t$.}
    \label{Fig:JFNK_Limitation}
\end{figure}

As illustrated in Figure  \ref{Fig:JFNK_Limitation}, we present two distinct scenarios: one where the JFNK method achieves successful convergence and another where it fails. In panel (a), the JFNK method converges successfully, with the relative residual descending to values proximate to zero. The log relative residual can be observed to decrease to values below -12. In contrast, Figure (b) showcases a situation where the JFNK method fails to converge. In such instances, the log relative residual stagnates at a particular value, preventing further reduction and resulting in non-convergence. Furthermore, by examining the phase space diagrams, it becomes evident that regardless of whether the convergence is successful or not, there exists a significant distance between the JFNK method's initial guess and the convergence point.

\section*{Appendix B: Fine-tuning of the DRL framework}

\subsection{Bayesian optimization of hyperparameters}

Bayesian optimization stands as a potent technique for optimizing intricate systems through an intelligent exploration and exploitation of the search space. This approach seamlessly integrates probabilistic modeling and optimization algorithms to construct an agent model that accurately captures the objective function's behavior \cite{BO1}. As a result, it enables informed decision-making while striking a balance between exploration and exploitation. By skillfully incorporating prior knowledge and effectively handling uncertainty, Bayesian optimization proves to be remarkably efficient in identifying global optima, even in scenarios marked by noise or limited data \cite{BO2}. 

Considering the pivotal role of two hyperparameters in our tasks— the external force distribution parameter $m$ and the standard deviation $\sigma$ — in modulating the reinforcement effect, we employ Bayesian optimization within our 2D KSE reinforcement learning framework. This approach facilitates a thorough investigation into the influence of these parameters on the reinforcement learning process, whilst concurrently ascertaining their optimal values for the 2D KSE \cite{BO3}. 

\begin{figure}
  \centering
  \includegraphics[width=1.0\textwidth]{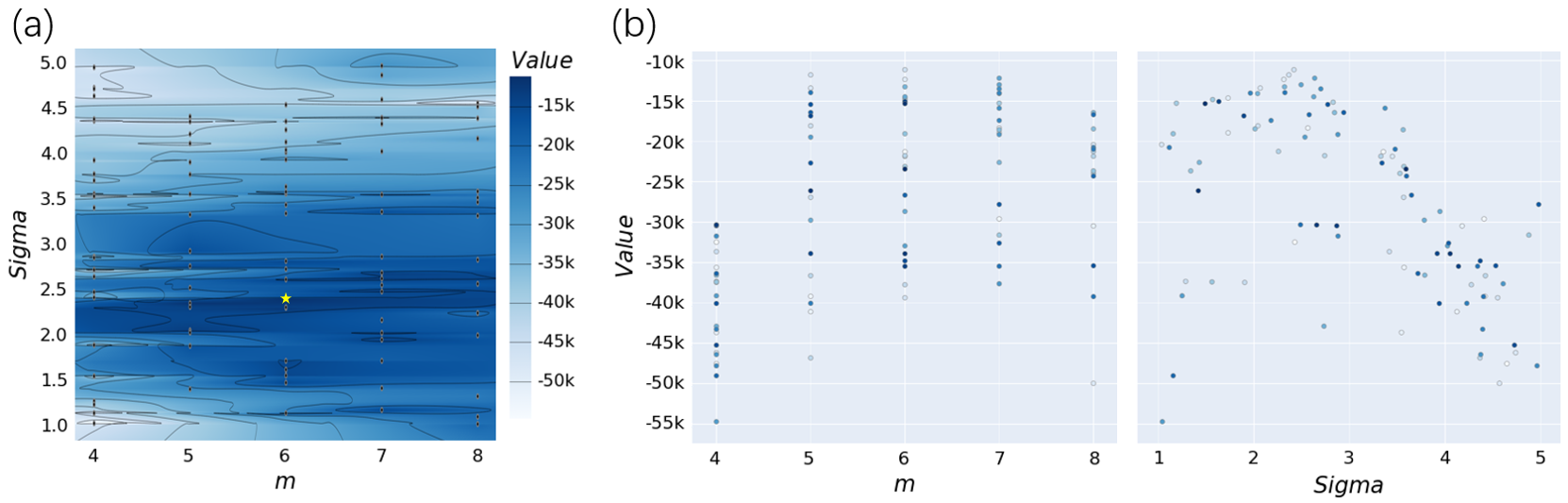}
  \caption{Bayesian optimization visualization: (a) contour plot and (b) slice plots. In the contour plot, the x-axis denotes the force distribution parameter $m$, while the y-axis signifies the standard deviation $\sigma$. Each point on the grid illustrates a distinct combination of the selected hyperparameters. Notably, the color or contour associated with each point is indicative of the objective value attained when employing that specific set of hyperparameters. The yellow pentagram denotes the location corresponding to the currently observed optimal objective values for $m$ and $\sigma$. In the slice plots, the x-axis corresponds to parameters $m$ and $\sigma$ respectively. The y-axis depicts the objective value. It is important to note that the color of each point serves solely as a distinction between different hyperparameter combinations and does not carry any physical meaning. }
    \label{Fig:BO}
  
\end{figure}

We employ the Bayesian optimization technique on the 2D KSE utilizing Python's Optuna Package and subsequently visualise the outcomes as depicted in Figure \ref{Fig:BO}. During the Bayesian Optimization procedure, while maintaining consistency in the initial and final points, we allow the DRL agent to undergo training under varying hyperparameters—distinct external force distributions parameter $m$ and standard deviation $\sigma$. Each training cycle was restricted to 300 episodes. At the termination of each episode, we accumulate the negative distance between the current system state and the goal point in each time step. This accumulation forms an objective value that serves as a reflection of the DRL agent's learning performance.

As illustrated in Figure \ref{Fig:BO}, concerning the hyperparameters $m$ and $\sigma$, we conduct a set of 100 Bayesian optimization experiments and subsequently plotted both contour and slice plots. From the slice plots, it is evident that both $m$ and $\sigma$ play crucial roles in the efficacy of reinforcement learning.  Moreover, observations from the contour plot indicate that when the external force distribution $m = 6$ and the standard deviation $\sigma$ $\approx 2.4$, the objective value is maximized. This result underscores that the DRL algorithm exhibits its optimal performance under these specific hyperparameters.

\subsection{Parallel reinforcement learning analysis}\label{parallel_section}

\begin{figure}
  \centering
  \includegraphics[width=1.0\textwidth]{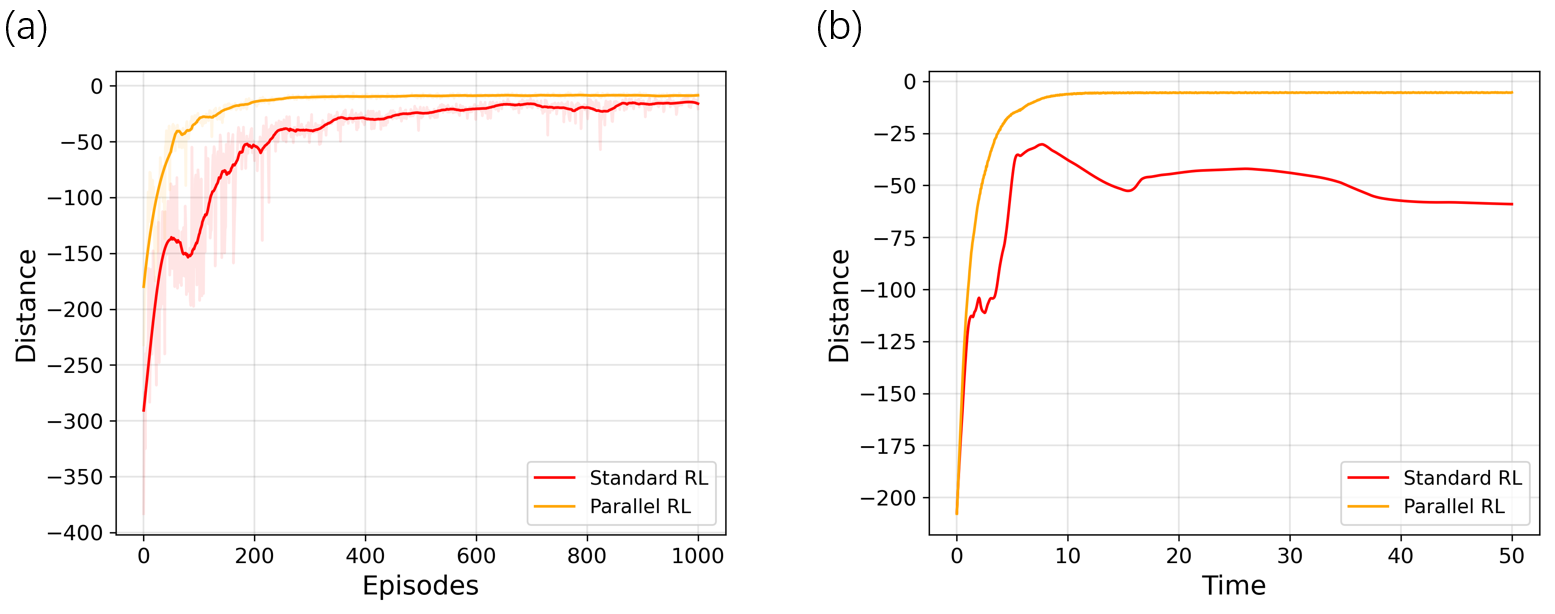}
\caption{Distance figure in (a) the DRL training process and (b) the test case for parallel DRL analysis. Compared with standard DRL, parallel DRL exhibits superior efficiency and stability throughout both training and testing phases.}
  \label{Fig:MultiEnvs}
\end{figure}

We also notice that the implementation of parallel reinforcement learning strategies significantly augments the stability and reliability of DRL systems. To elaborate, in our research, we deploy ten agents to concurrently accumulate learning trajectories. These agents share the same neural network, thereby adhering to a uniform DRL policy. The data amassed by these ten agents is collectively added to the replay buffer after each time step, which is then utilized to update the policy of the DRL agents. 

Although parallel reinforcement learning is employed in both tasks (i.e., navigating towards a fixed goal point and identifying fixed points), considering the high level of randomness in the task of identifying fixed points, we use the scenario of navigating towards a goal fixed point below to illustrate the effectiveness of parallel reinforcement learning.
As depicted in Figure \ref{Fig:MultiEnvs}, a comparison between the direct employment of reinforcement learning and its parallel variant reveals distinct advantages associated with parallel reinforcement learning. First and foremost, the parallel approach demonstrates marked improvements in learning outcomes. Although standard DRL reduces the performance differential with increased training episodes, parallel DRL demonstrably excels in effectiveness and stability in practical test cases. On the other hand, parallel DRL exhibits a significant training speed advantage over standard DRL, achieving convergence in just 200 episodes. 

\subsection{Exploration noise optimization analysis}\label{noise_section}

\begin{figure}
  \centering
  \includegraphics[width=1.0\textwidth]{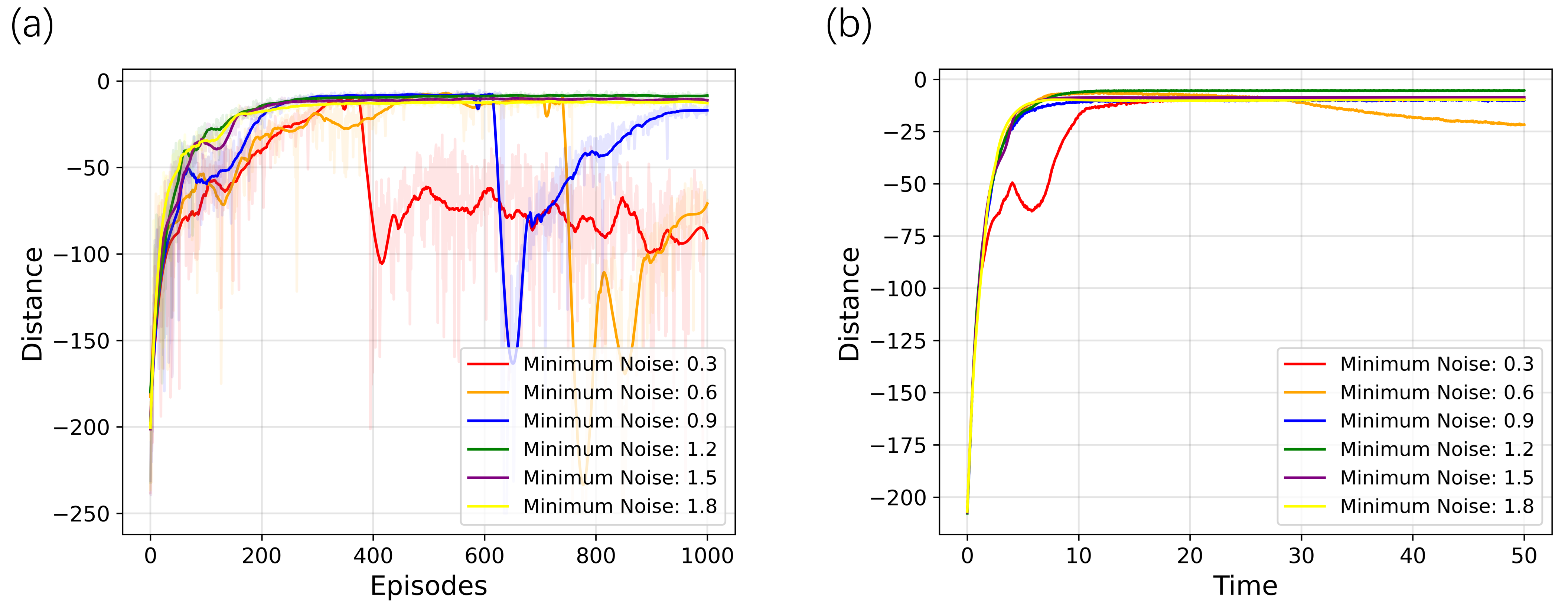}
  \caption{Distance in (a) the DRL training process and (b) the test case of different minimum exploration noise parameter $\alpha_{min}$. It is evident that the optimal performance is achieved when $\alpha_{min} = 1.2$. }
    \label{Fig:Noise}
\end{figure}

As delineated in Section \ref{method}\ref{noise}, the minimum exploration noise parameter, denoted as $\alpha_{min}$, assumes a pivotal role in our reinforcement learning algorithm. This parameter serves as a fundamental element in harmonizing the dichotomy between exploration and exploitation within the DRL frameworks. Consequently, an in-depth analysis of the minimum exploration noise parameter $\alpha_{min}$ has been conducted. 


Figure \ref{Fig:Noise} presents the effect of $\alpha_{min}$, in optimizing the effectiveness of the DRL processes in the task of navigating towards a goal fixed point.
The empirical evidence suggests that a setting of $\alpha_{min}=1.2$ is optimal, as it leads to superior learning performance in the DRL agent. 
Upon analyzing the data in the figure for the training process, we discern that a lower threshold ($\alpha_{min} < 0.9$) results in the DRL agent's predominant exploitation of existing knowledge and potential entrapment in local optima. Due to the complexity of 2D KSE environment and overreliance on known strategies, DRL agent may be unable to handle diverse complex case in training process, leading to significant performance fluctuations and inefficient learning. On the other hand, a higher threshold ($\alpha_{min} > 1.5$) induces excessive exploration, whereby the agent may prematurely abandon efficient strategies in the pursuit of potentially better alternatives, slightly diminishing the effectiveness of DRL agent. In Figure \ref{Fig:Noise}, the variations in effectiveness across different $\alpha_{min}$  settings are relatively minor. Nonetheless, it is evident that the $\alpha_{min} = 1.2$ produces the optimal result. 

We would like to highlight that, owing to the variance in scenarios, the minimum exploration noise parameter implemented in the 'identifying fixed points' task exhibits some differences compared to that used in the 'navigating towards a goal fixed point' task. In the task of identifying fixed points, we employed the same method for the optimization of the minimum exploration noise parameter, achieving the optimal result at $\alpha_{min} = 1.5$.

\bibliographystyle{RS} 
\bibliography{fixedpoint_2DKSE_revision} 

\end{document}